%% file: naacl_2025.tex
\pdfoutput=1

\documentclass[11pt]{article}

\usepackage[final]{acl}

\usepackage{times}
\usepackage{latexsym}

\usepackage[T1]{fontenc}

\usepackage[utf8]{inputenc}
\usepackage{float}
\usepackage{booktabs}  
\usepackage{amsmath}  
\usepackage{microtype}
\usepackage{multirow}

\usepackage{inconsolata}

\usepackage{graphicx}
\usepackage{listings}
\usepackage{subcaption}
\usepackage{multirow}
\usepackage{booktabs}
\usepackage{array}
\usepackage{makecell}
\usepackage{graphicx}
\usepackage{geometry}
\usepackage{tikz}
\usepackage{booktabs}

\usepackage{geometry}
\usepackage{tikz}
\usepackage{booktabs}

\title{Rationale Behind Essay Scores: Enhancing S-LLM's Multi-Trait Essay Scoring with Rationale Generated by LLMs}

\author{
  Seong Yeub Chu\textsuperscript{1}\thanks{Both authors contributed equally to this research.}, 
  Jong Woo Kim\textsuperscript{2}\footnotemark[1], 
  Bryan Wong\textsuperscript{1}, 
  Mun Yong Yi\textsuperscript{1}\thanks{Corresponding author.} \\
  \textsuperscript{1}Graduate School of Data Science, KAIST \\
  \textsuperscript{2}Department of Industrial \& Systems Engineering, KAIST \\
  \texttt{\{chseye7, gsds4885, bryan.wong, munyi\}@kaist.ac.kr}
}

\begin{document}
\maketitle

\begin{abstract}
    \input{abstract}

\end{abstract}

\section{Introduction}
\label{intro}
\input{intro}

\section{Related Work}
\label{related_work}
\input{related_work}

\section{RMTS}
\label{rmts}
\input{rmts}

\section{Experiment}
\label{experiment}
\input{experiment}

\section{Results}
\label{results}
\input{results}

\section{Conclusion}
\label{conclusion}
\input{conclusion}

\section*{Limitations}
In this study, we have identified two primary limitations. First, our model's performance could be affected by the sequence order of traits due to the use of autoregressive models like T5 and BART. Future research should explore models like XLNet \cite{yang_xlnet}, which are better suited for handling sequence orders. Secondly, we focused exclusively on multi-trait scoring of English writing. To evaluate the scalability of our model for general language education, further studies are needed on languages other than English.

\section*{Ethical Statement}
This study utilizes only publicly available benchmark datasets, including ASAP, ASAP++, and Feedback Prize. 

\section*{Acknowledgements}
This research was supported by Chunjae Education Inc. (Project Number G01250064).

\bibliography{ref}
\newpage
\newpage
\section*{Appendix}
\appendix

\section{Details of the baselines}
\label{baseline_detail}

\begin{itemize}
    \item HISK \cite{cozma}: is a string kernel based on histogram intersection, used in combination with a support vector regressor.
    \item STL-LSTM \cite{dong2017}\footnote{https://github.com/feidong1991/aes.git}: uses a combination of LSTM and CNN to infer essay scores of every trait individually.
    \item MTL-BiLSTM \cite{kumar}\footnote{https://github.com/ASAP-AEG/MTL-Essay-Traits-Scoring.git}: employs trait-specific BiLSTM layers to score multi-trait, ultimately predicting the overall score.
    \item R2BERT \cite{yang2020enhancing}: enhances AES by fine-tuning pre-trained language models, combining regression and ranking objectives to improve scoring accuracy. 
    \item NPCR \cite{xie2022automated}\footnote{https://github.com/CarryCKW/AES-NPCR.git}: employs pairwise contrastive regression to learn relative score differences between essays, thereby refining the scoring process.
    \item PMAES \cite{chen2023pmaes}\footnote{https://github.com/gdufsnlp/PMAES.git}: utilizes prompt-mapping contrastive learning to generalize scoring across various prompts, enhancing the model's adaptability.
    \item PLAES \cite{chen2024plaes}: introduces a prompt-generalized and level-aware learning framework, improving cross-prompt AES performance by considering prompt variations and essay complexity levels.
    \item T5\footnote{https://huggingface.co/google-t5/t5-base} \cite{raffel}: is a transformer-based model that frames NLP tasks as a text-to-text problem. In this study, we used the "google-t5/t5-base" model.
    \item Flan-T5\footnote{https://huggingface.co/google/flan-t5-base} \cite{flant5}: builds upon T5 \cite{raffel} by introducing fine-tuning on instruction-based datasets. In this study, we used the "google/flan-t5-base" model.
    \item BART\footnote{https://huggingface.co/facebook/bart-base} \cite{lewis}: is a sequence-to-sequence models trained by corrupting text and learning to reconstruct the original text. In this study, we used the "facebook/bart-base" model.
    \item Pegasus\footnote{https://huggingface.co/google/pegasus-x-base} \cite{pegasus}: is designed specifically for abstractive summarization tasks, focusing on predicting whole sentences that have been masked. In this study, we used "google/pegasus-x-base" model.
    \item LED\footnote{https://huggingface.co/allenai/led-large-16384} \cite{led}: extends the transformer architecture to handle longer documents efficiently by using sparse attention mechanisms. In this study, we used "allenai/led-large-16384" model.
\end{itemize}

\section{Length Statistics of Rationales}
\label{length_statistics}
\begin{figure}[h!]
    \centering
    \begin{subfigure}[b]{\columnwidth} 
        \centering
        \includegraphics[width=0.9\textwidth]{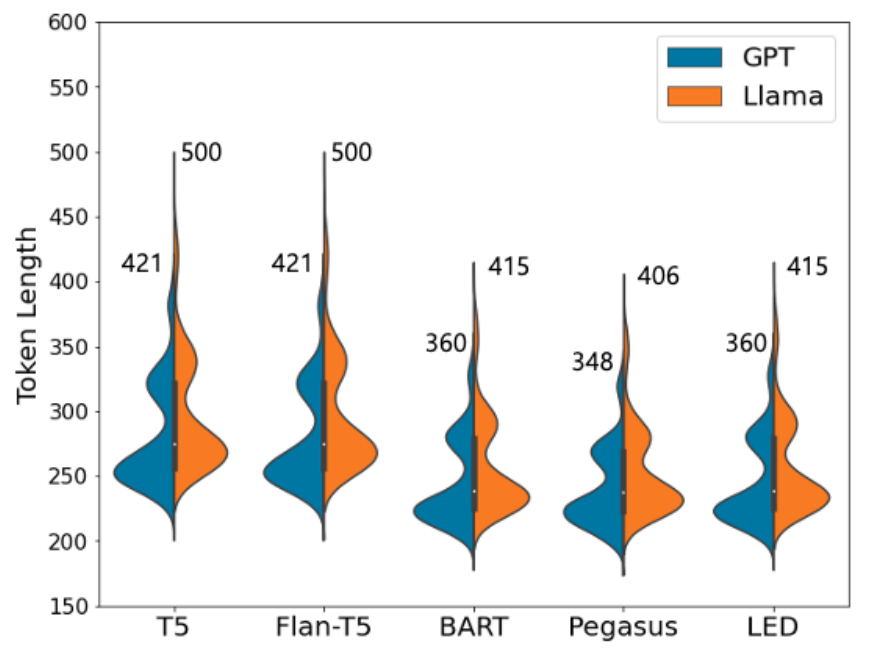} 
        \caption{ASAP/ASAP++}
    \end{subfigure}
    
    \vspace{-0cm} 
    
    \begin{subfigure}[b]{\columnwidth} 
        \centering
        \includegraphics[width=0.9\textwidth ]{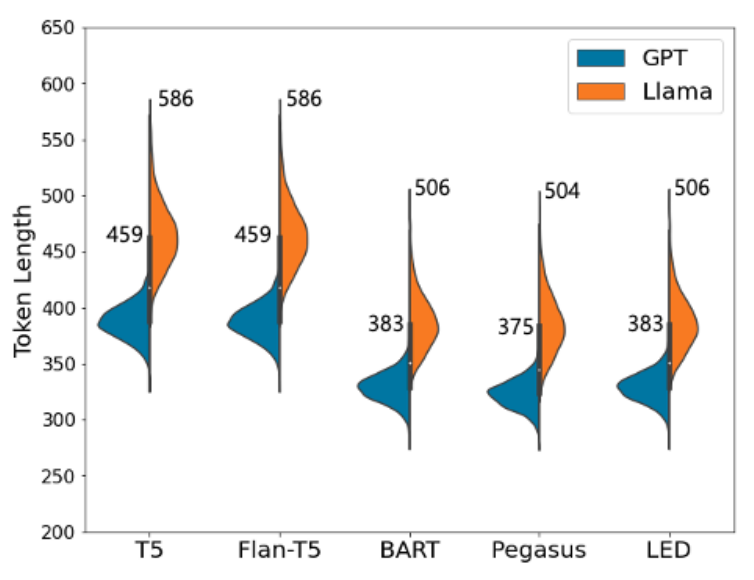} 
        \caption{Feedback}
    \end{subfigure}
    
    \caption{Violin plots of the number of tokens per rationale depending on each individual model's tokenizer. (A) refers to the rationale from the ASAP/ASAP++ dataset and (B) refers to the rationale from the Feedback Prize dataset.}
    \label{token_length_plot}
\end{figure}

The generated rationales were tokenized using each model's corresponding tokenizer. As shown in Figure \ref{token_length_plot}, aside from the rationales from the Feedback dataset generated by Llama (which reached a maximum of 586 tokens when tokenized by the T5 tokenizer), the maximum number of tokens in the rationales produced by either GPT or Llama did not exceed 512. This is the typical limit that transformer-based language models can process. This suggests that the rationale lengths are manageable and should not impede the models' ability to capture contextual information. For Llama-generated rationales in the Feedback Prize dataset, any rationales exceeding 512 tokens were truncated to comply with the limit. Interestingly, Llama tended to generate longer rationales than GPT in the Feedback dataset.

\section{Faithfulness of Rationales from ASAP/ASAP++ and Feedback Prize Datasets} 
\label{appendix_faithfulness}

This section presents the faithfulness analysis of rationales generated by LLMs on the ASAP/ASAP++ and Feedback Prize datasets. We compare model performance using Quadratic Weighted Kappa (QWK) scores when scoring essays versus LLM-generated rationales.

\subsection{ASAP/ASAP++ Dataset}
\begin{figure}[h!]
    \centering
    \includegraphics[trim=0 0 0 0, clip, width=0.8\linewidth]{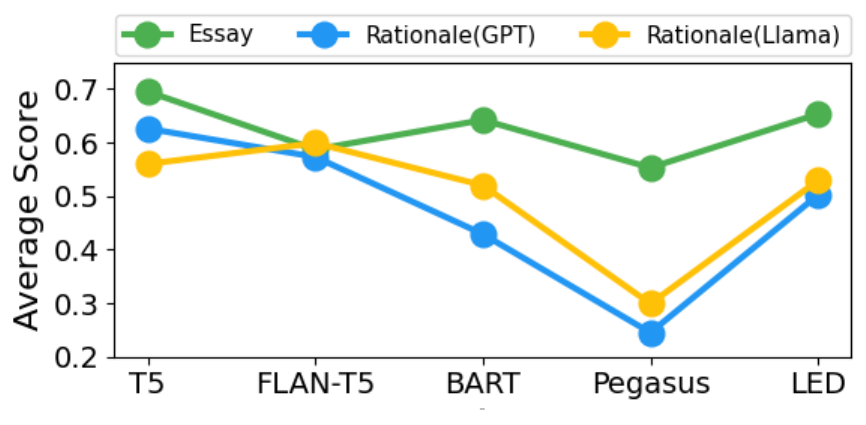}
    \caption{Performance comparison based on QWK scores, averaged across all traits, for the ASAP/ASAP++ dataset, using either the essays or the rationales generated by GPT or Llama.}
    \label{faithfulness_fig}
\end{figure}

Figure \ref{faithfulness_fig} shows the QWK performance of five S-LLMs on the ASAP/ASAP++ dataset using only LLM-generated rationales.

\subsection{Feedback Prize Dataset}
\begin{figure}[H]
    \centering
    \includegraphics[width=\linewidth]{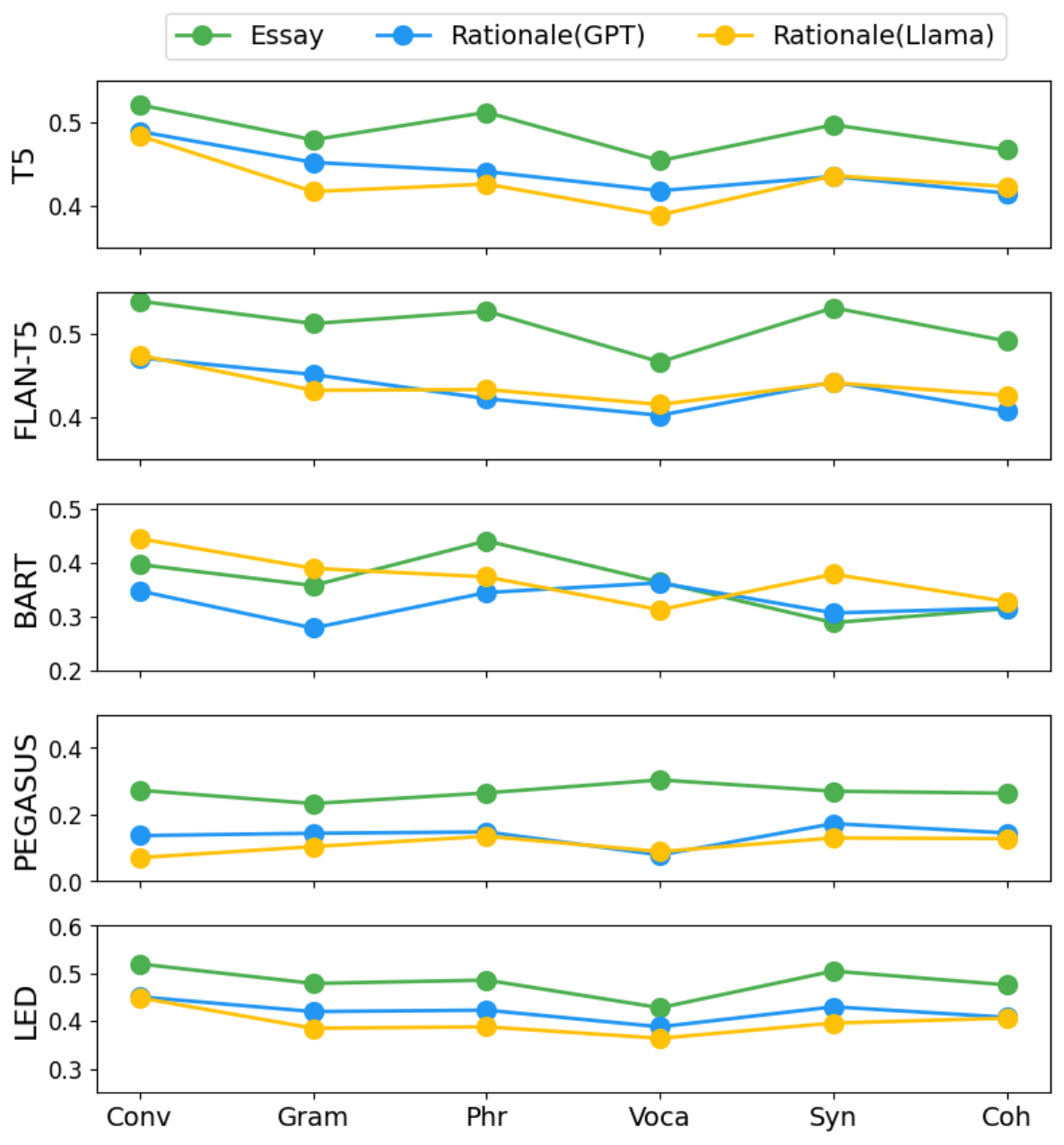}
    \caption{Performance comparison of each model based on QWK scores, averaged across all prompts for each trait with regard to the Feedback Prize dataset, using either the essays or the rationales generated by GPT or Llama.}
    \label{essay_rationale_trait_feedback}
\end{figure}

\begin{figure}[H]
    \centering
    \includegraphics[width=0.8\linewidth]{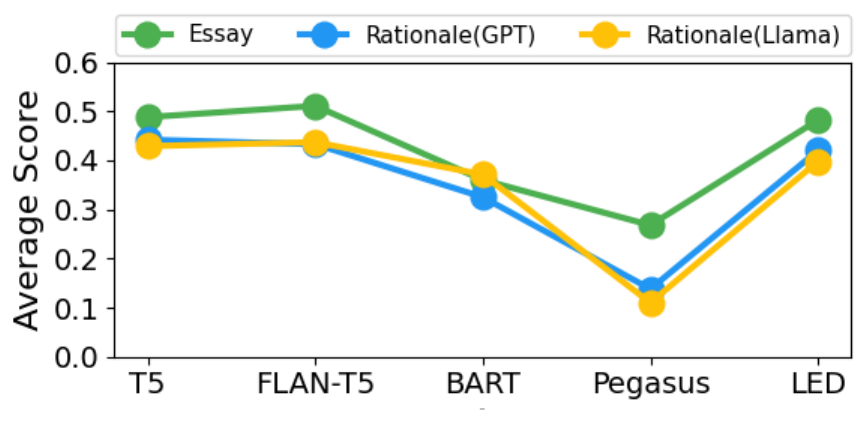}
    \caption{Performance comparison based on QWK scores, averaged across all traits, with regard to the Feedback Prize dataset using either the essays or the rationales generated by GPT or Llama.}
    \label{essay_rationale_feedback}
\end{figure}

Figures \ref{essay_rationale_feedback} and \ref{essay_rationale_trait_feedback} illustrate the model performance in QWK on the Feedback Prize dataset using only LLM-generated rationales.

\begin{table*}[]
    \centering
    \caption{Performance of Compressed ASAP Rationale. The models are abbreviated as follows: G = GPT-based original rationale, G+C = GPT-based rationale with compression, L = Llama-based original rationale, L+C = Llama-based rationale with compression.}
    \resizebox{\textwidth}{!}{%
    \begin{tabular}{lcccccccccccc}
        \toprule
        Model & Overall & Con & PA & Lang & Nar & Org & Conv & WC & SF & Style & Voice & AVG \\
        \midrule
        T5 RMTS(G) & 0.755 & 0.737 & 0.752 & 0.713 & 0.744 & 0.682 & 0.690 & 0.705 & 0.694 & 0.702 & 0.612 & 0.708 \\
        T5 RMTS(G+C) & 0.754 & 0.733 & 0.757 & 0.703 & 0.740 & 0.680 & 0.696 & 0.691 & 0.688 & 0.697 & 0.604 & 0.704 \\
        T5 RMTS(L) & 0.754 & 0.730 & 0.749 & 0.701 & 0.737 & 0.675 & 0.684 & 0.690 & 0.684 & 0.696 & 0.640 & 0.704 \\
        T5 RMTS(L+C) & 0.752 & 0.722 & 0.746 & 0.694 & 0.734 & 0.672 & 0.679 & 0.682 & 0.682 & 0.692 & 0.621 & 0.698 \\
        \bottomrule
    \end{tabular}%
    }
    \label{tab:asap_detailed}
\end{table*}

\begin{table}
    \centering
    \footnotesize
    \caption{Performance of Compressed Feedback Rationale. 
    The models are abbreviated as follows: G = GPT-based, G+C = GPT-based with Compression, L = Llama-based, L+C = Llama-based with Compression.}
    \resizebox{\columnwidth}{!}{%
    \begin{tabular}{lccccccc}
        \toprule
        Model & Conv & Gram & Phr & Voc & Syn & Coh & Avg \\
        \midrule
        T5 RMTS (G) & 0.568 & 0.550 & 0.543 & 0.430 & 0.543 & 0.498 & 0.522 \\
        T5 RMTS (G+C) & 0.521 & 0.479 & 0.512 & 0.414 & 0.497 & 0.467 & 0.482 \\
        T5 RMTS (L) & 570 & 0.557 & 0.535 & 0.443 & 0.534 & 0.490 & 0.522 \\
        T5 RMTS (L+C) & 0.553 & 0.521 & 0.527 & 0.412 & 0.515 & 0.479 & 0.501 \\
        \bottomrule
    \end{tabular}%
    }
    \label{tab:feedback_additional}
\end{table}

\section{Additional Experiments}
\subsection{Result of Compressed Rationales}

As shown in Appendix \ref{length_statistics}, some rationales were relatively long and could have hindered S-LLMs' scoring performance. To address this, we designed a prompt to guide the LLM in removing redundant information, making the rationales more concise while retaining key details.

An analysis of the ASAP dataset showed that the average rationale length was significantly higher for the original rationales than for the shortened versions across all tested models. For instance, in the GPT-T5 setting, the original rationales averaged 277.55 words, with a maximum of 431 words, whereas the compressed versions averaged 133.04 words, with a maximum of 228 words. Similarly, in the Llama-T5 setting, original rationales averaged 298.62 words (max 500), compared to 143.35 words (max 238) for the shortened versions.

To assess the impact of rationale length, we compared T5’s performance using original versus shortened rationales. As shown in Table \ref{tab:asap_detailed} and Table \ref{tab:feedback_additional}, models using original rationales generally outperformed those using compressed versions on average. This suggests that the original rationales provided richer information that improved the model’s ability to score essays reliably.

\subsection{Comparison of \textit{Overall} Score Prediction}
\label{overall_comparison}

While RMTS is inherently designed for multi-trait scoring, we adapted it to generate a single \textit{overall} score for direct comparison with models that do not support multi-trait scoring.

Table \ref{tab:overall_comparison} shows the QWK performance of our model compared to two baselines designed for \textit{overall} score prediction: R2Bert \cite{yang2020enhancing} and NPCR \cite{xie2022automated}. T5 RMTS (GPT-3.5) achieves the highest average QWK score of 0.758, outperforming NPCR, the best-performing baseline, by a significant margin (7.7\%).

Additionally, RMTS maintains strong performance across different prompts, demonstrating its generalization ability. These results confirm that RMTS excels not only in multi-trait scoring but also in \textit{overall} score prediction.

\begin{table}[H]
    \centering
    \caption{Overall QWK Score Comparison Across Models}
    \resizebox{\columnwidth}{!}{%
    \begin{tabular}{lccccccccc}
        \toprule
        Model & 1 & 2 & 3 & 4 & 5 & 6 & 7 & 8 & AVG \\
        \midrule
        R2Bert & 0.627 & 0.533 & 0.517 & 0.757 & 0.610 & 0.685 & 0.683 & 0.428 & 0.605 \\
        NPCR & 0.719 & 0.586 & 0.650 & 0.751 & 0.766 & 0.753 & 0.683 & 0.543 & 0.681 \\
        \midrule
        \textbf{T5 RMTS(G)} & \textbf{0.768} & \textbf{0.655} & \textbf{0.712} & \textbf{0.810} & \textbf{0.810} & \textbf{0.808} & \textbf{0.792} & \textbf{0.710} & \textbf{0.758} \\
        \bottomrule
    \end{tabular}%
    }
    \label{tab:overall_comparison}
\end{table}

\subsection{Ablation study of Feedback Prize Dataset }

\begin{figure}[H]
    \centering
    \includegraphics[width=\linewidth]{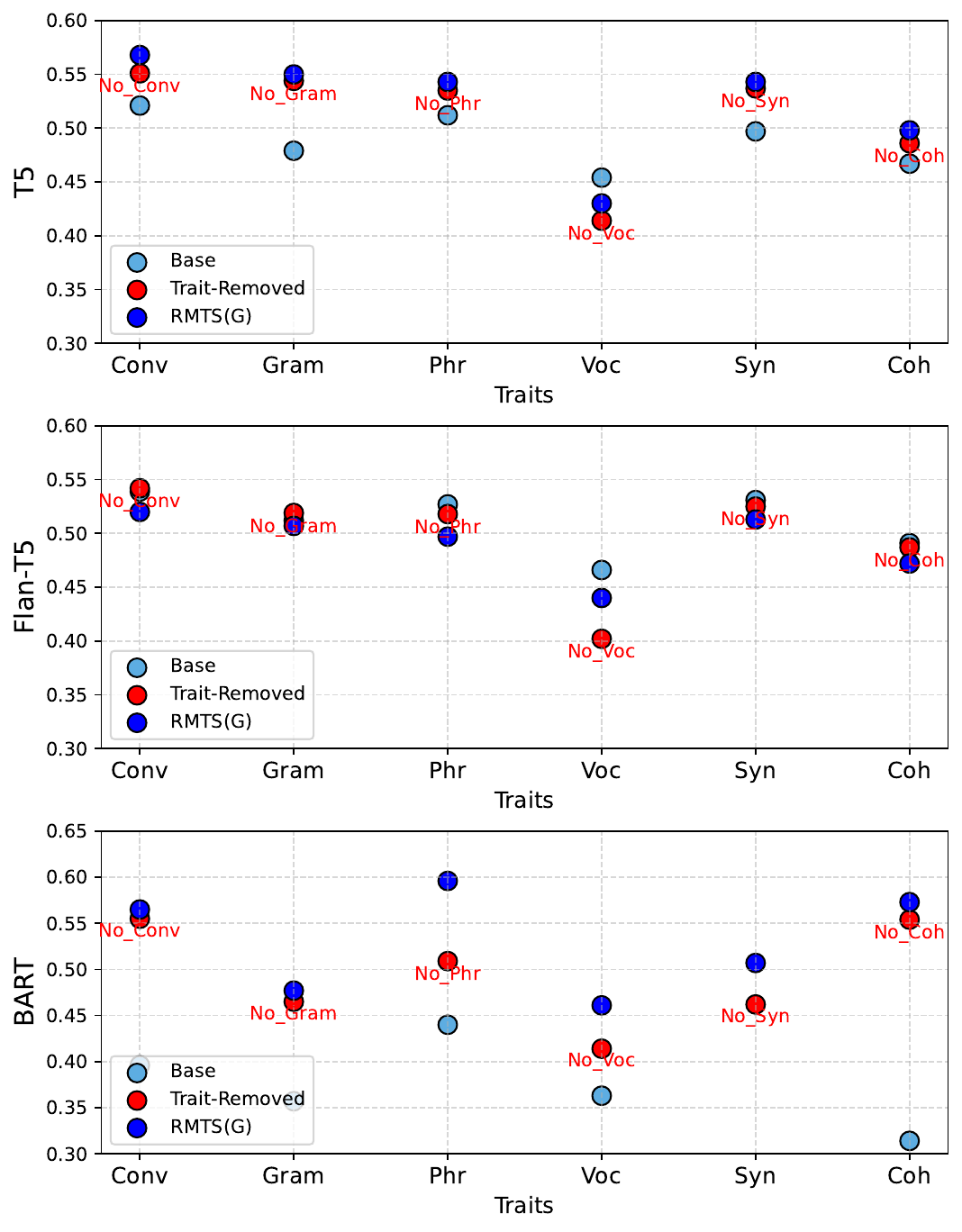}
    \caption{Ablation study on the Feedback dataset when removing each trait's rationale.}
    \label{fig:enter-label}
\end{figure}

To assess the impact of trait-specific rationales, we conducted an additional experiment on the Feedback dataset (Figure \ref{fig:enter-label}). Using T5, Flan-T5, and BART, we fine-tuned each model following the same procedure described in Section \ref{asap_ablation_section}. The results show a consistent performance drop for a given trait when its corresponding rationale is removed, confirming its importance in trait-specific assessment. Notably, models without a specific trait rationale still outperform baseline models without rationales, suggesting that rationales contribute beyond their assigned trait evaluation. However, in some cases, removing a trait rationale improved scoring performance on Flan-T5, indicating that rationale integration does not enhance its performance on the Feedback dataset, likely due to its small size, as discussed in Section \ref{feedback_performance_section}. Although the extent of performance degradation varies across traits and models, the overall trend confirms that trait-specific rationales improve scoring accuracy, as their removal generally weakens performance.

\section{LLM Settings}
For RMTS, we used GPT-3.5-Turbo and LLama3.1-8B-Instruct provided by OpenAI and Meta. GPT-3.5-Turbo was used in the form of API\footnote{https://openai.com/index/openai-api/}, and LLama 3.1-8B-Instruct was employed by utilizing the official code shared by Meta\footnote{https://llama.meta.com/responsible-use-guide/}. Regarding GPT, we performed the experiments with \textit{gpt-3.5-turbo-0125}. When this study was conducted, the cost for processing input tokens with the model was \$0.5 per 1M tokens, while generating output tokens was priced at \$1.5 per 1M tokens. We consistently used identical hyperparameters: a temperature of 0, frequency and presence penalties both set to 0, and a Top-p value of 1 for the cumulative probability cutoff in nucleus sampling. Given that the temperature hyperparameter is set to 0, we conducted the experiment a single time. For prompts 3 to 6 of ASAP++, excerpts were excluded in both LLMs.

\section{Prompts and System Message}
Examples of prompts and system messages used by the LLMs to generate rationales can be found in Appendix \ref{appendix_f}. We revised and supplemented \cite{lee} by adding trait-specific rubric and additional prompts to generate rationales, designing a comprehensive prompt template.

\newpage 

\onecolumn

\section{Examples of System Messages and Predefined Template} \label{appendix_f}

\subsection{System Message}
The system message corresponding to each agent used in our experiment are as follows.

\noindent\textbf{System message:} You are a member of the English essay writing test evaluation committee. Please, evaluate given essay using following information.

\subsection{Predefined Template (\textit{Prompt 1}, \textit{Content})}
\textbf{[Prompt]}\\ 
More and more people use computers, but not everyone agrees that this benefits society. Those
who support advances in technology believe that computers have a positive effect on people.
They teach hand-eye coordination, give people the ability to learn about faraway places and
people, and even allow people to talk online with other people. Others have different ideas.
Some experts are concerned that people are spending too much time on their computers and
less time exercising, enjoying nature, and interacting with family and friends.
Write a letter to your local newspaper in which you state your opinion on the effects
computers have on people. Persuade the readers to agree with you.\\
\textbf{(end of [Prompt])}

\noindent\textbf{[Trait-Specific Rubric Guidelines]}\\
This property checks for the amount of content and ideas present in the essay.

\noindent Score 6: The writing is exceptionally clear, focused, and interesting. It holds the reader’s
attention throughout. Main ideas stand out and are developed by strong support and rich details
suitable to audience and purpose. The writing is characterized by\\
\noindent • clarity, focus, and control.  \\
\noindent • main idea(s) that stand out.\\
\noindent • supporting, relevant, carefully selected details; when appropriate, use of
resources provides strong, accurate, credible support.\\
\noindent • a thorough, balanced, in-depth explanation / exploration of the topic; the writing
makes connections and shares insights.\\
\noindent • content and selected details that are well-suited to audience and purpose.\\

\noindent Score 5: The writing is clear, focused and interesting. It holds the reader’s attention. Main ideas
stand out and are developed by supporting details suitable to audience and purpose. The writing is characterized by\\
\noindent • clarity, focus, and control.\\
\noindent • main idea(s) that stand out.\\
\noindent • supporting, relevant, carefully selected details; when appropriate, use of resources provides strong, accurate, credible support.\\
\noindent • a thorough, balanced explanation / exploration of the topic; the writing makes connections and shares insights.\\
\noindent • content and selected details that are well-suited to audience and purpose.\\

\noindent Score 4: The writing is clear and focused. The reader can easily understand the main ideas. Support is present, although it may be limited or rather general. The writing is characterized by\\
\noindent • an easily identifiable purpose.\\
\noindent • clear main idea(s).\\
\noindent • supporting details that are relevant, but may be overly general or limited in places; when appropriate, resources are used to provide accurate support.\\
\noindent • a topic that is explored / explained, although developmental details may occasionally be out of balance with the main idea(s); some connections and insights may be present.\\
\noindent • content and selected details that are relevant, but perhaps not consistently well-chosen for audience and purpose.\\

\noindent Score 3: The reader can understand the main ideas, although they may be overly broad or simplistic, and the results may not be effective. Supporting detail is often limited, insubstantial, overly general, or occasionally slightly off-topic. The writing is characterized by\\
\noindent • an easily identifiable purpose and main idea(s).\\
\noindent • predictable or overly-obvious main ideas; or points that echo observations heard elsewhere; or a close retelling of another work.\\
\noindent • support that is attempted, but developmental details are often limited, uneven, somewhat off-topic, predictable, or too general (e.g., a list of underdeveloped points).\\
\noindent • details that may not be well-grounded in credible resources; they may be based on clichés, stereotypes or questionable sources of information.\\
\noindent • difficulties when moving from general observations to specifics.\\

\noindent Score 2: Main ideas and purpose are somewhat unclear or development is attempted but minimal. The writing is characterized by\\
\noindent • a purpose and main idea(s) that may require extensive inferences by the reader.\\
\noindent • minimal development; insufficient details.\\
\noindent • irrelevant details that clutter the text.\\
\noindent • extensive repetition of detail.\\

\noindent Score 1: The writing lacks a central idea or purpose. The writing is characterized by\\
\noindent • ideas that are extremely limited or simply unclear.\\
\noindent • attempts at development that are minimal or nonexistent; the paper is too short to demonstrate the development of an idea.\\
\textbf{(end of [Trait-Specific Rubric Guidelines])}\\

Refer to the provided [Prompt], and [Trait-Specific Rubric Guidelines] to evaluate the given essay.\\

\noindent\textbf{[Note]}\\
I have made an effort to remove personally identifying information from the essays using the Named Entity Recognizer (NER). The relevant entities are identified in the text and then replaced with a string
such as "@PERSON", "@ORGANIZATION", "@LOCATION", "@DATE", "@TIME", "@MONEY", "@PERCENT", "@CAPS" (any capitalized word) and "@NUM" (any digits). Please do not penalize the essay because of the anonymizations.\\
\textbf{(end of [Note])}\\

\noindent\textbf{[Essay]}\\
\{essay\}\\
\textbf{(end of [Essay])}\\

\noindent Q. List the quotations from the [Essay] that are relevant to "content" and evaluate whether each quotation is well-written or not.
Provide a literal evaluation of the [Essay], returning text-based feedback rather than scores. Refrain from offering additional commentary.
Please keep your response within 50 words.

\section{Examples of Rationales}
\subsection{A Rationale of an Essay for Prompt 1 of ASAP++}
\begin{itemize}
    \item \textbf{Essay:} Dear @CAPS1 @CAPS2, I believe that using computers will benefit us in many ways like talking and becoming friends will others through websites like facebook and mysace. Using computers can help us find coordibates, locations, and able ourselfs to millions of information. Also computers will benefit us by helping with jobs as in planning a house plan and typing a @NUM1 page report for one of our jobs in less than writing it. Now lets go into the wonder world of technology. Using a computer will help us in life by talking or making friends on line. Many people have myspace, facebooks, aim, these all benefit us by having conversations with one another. Many people believe computers are bad but how can you make friends if you can never talk to them? I am very fortunate for having a computer that can help with not only school work but my social life and how I make friends. Computers help us with finding our locations, coordibates and millions of information online. If we didn't go on the internet a lot we wouldn't know how to go onto websites that @MONTH1 help us with locations and coordinates like @LOCATION1. Would you rather use a computer or be in @LOCATION3. When your supposed to be vacationing in @LOCATION2. Million of information is found on the internet. You can as almost every question and a computer will have it. Would you rather easily draw up a house plan on the computers or take @NUM1 hours doing one by hand with ugly erazer marks all over it, you are garrenteed that to find a job with a drawing like that. Also when appling for a job many workers must write very long papers like a @NUM3 word essay on why this job fits you the most, and many people I know don't like writing @NUM3 words non-stopp for hours when it could take them I hav an a computer. That is why computers we needed a lot now adays. I hope this essay has impacted your descion on computers because they are great machines to work with. The other day I showed my mom how to use a computer and she said it was the greatest invention sense sliced bread! Now go out and buy a computer to help you chat online with friends, find locations and millions of information on one click of the button and help your self with getting a job with neat, prepared, printed work that your boss will love.\\
    \item \textbf{Ground truth trait scores:} \textit{Content: 4/6, Organization: 4/6, Word Choice: 4/6, Sentence Fluency: 3/6, Conventions: 4/6})\\
    \item \textbf{Rationale (GPT-3.5-Turbo):} \textbf{[sentence fluency]:} The essay lacks sentence fluency due to inconsistent structure, awkward phrasing, and errors in grammar and punctuation. The ideas are presented in a disjointed manner, making it challenging for the reader to follow smoothly. The essay would benefit from improved sentence variety and coherence.\textbf{[word choice]:} The essay lacks precision and variety in word choice. There are numerous spelling and grammar errors, misuse of words, and awkward phrasing. The vocabulary is limited, and the language is not engaging or effective in persuading the audience. The essay would benefit from significant revisions for clarity and impact.\textbf{[conventions]:} The essay demonstrates significant issues with standard writing conventions. There are frequent errors in punctuation, spelling, grammar, and capitalization, which impede readability and clarity. The lack of control over basic conventions hinders effective communication and requires substantial editing for improvement. \textbf{[organization]:} The essay lacks clear organization. Ideas are scattered and not well-developed, making it difficult to follow a logical flow. The structure of a letter to the editor is attempted but not effectively maintained. Improvement is needed in organizing paragraphs and connecting ideas cohesively. \textbf{[content]:} The essay discusses the benefits of using computers, such as online communication, access to information, and assistance with tasks like job applications. However, the ideas lack depth and coherence, with limited development and organization, impacting the overall clarity and effectiveness of the argument.\\
    \item \textbf{Rationale (Llama-3.1-8B-Instruct):}  \textbf{[sentence fluency]}: The essay has a choppy flow with abrupt sentence transitions and inconsistent sentence structure. Many sentences are short and lack variety, resulting in a monotonous rhythm. The use of fragments and run-on sentences further disrupts the flow, making it difficult to follow at times. \textbf{[word choice]}: The essay demonstrates a limited vocabulary, relying heavily on simple and generic words. The writer uses colloquial expressions and slang ("garrenteed," "I hav an") and misuses words ("coordibates," "able ourselfs"). The text also features overused phrases and clichés ("the greatest invention since sliced bread"). \textbf{[conventions]}: The essay demonstrates limited control of standard writing conventions, with numerous errors in punctuation, spelling, capitalization, and grammar. Internal punctuation is frequently incorrect, and spelling errors, such as "coordibates" and "mysace," significantly impede readability. \textbf{[organization]}: The essay lacks a clear and logical structure. Ideas are presented in a disjointed manner, jumping between topics such as the benefits of computers, social media, and job applications. Transitions between ideas are absent, making the essay difficult to follow. \textbf{[content]}: The essay lacks depth and balance, relying heavily on general statements and clichés. It fails to provide nuanced arguments, instead using simplistic and repetitive points to support its claim that computers benefit society. Relevant details are often overly general or limited.

\end{itemize}
\subsection{A Rationale of an Essay for Prompt 3 of Feedback Prize}
\begin{itemize}
    \item \textbf{Essay:} I think that students would benefit from learning at home,because they wont have to change and get up early in the morning to shower and do there hair. taking only classes helps them because at there house they'll be pay more attention. they will be comfortable at home. The hardest part of school is getting ready. you wake up go brush your teeth and go to your closet and look at your cloths. after you think you picked a outfit u go look in the mirror and youll either not like it or you look and see a stain. Then you'll have to change. with the online classes you can wear anything and stay home and you wont need to stress about what to wear.most students usually take showers before school. they either take it before they sleep or when they wake up. some students do both to smell good. that causes them do miss the bus and effects on there lesson time cause they come late to school. when u have online classes u wont need to miss lessons cause you can get everything set up and go take a shower and when u get out your ready to go.when your home your comfortable and you pay attention. it gives then an advantage to be smarter and even pass there classmates on class work. public schools are difficult even if you try. some teacher dont know how to teach it in then way that students understand it. that causes students to fail and they may repeat the class.\\
    \item \textbf{Ground Truth Trait Scores:} \textit{Cohesion: 3.5/5.0, Syntax: 3.5/5.0, Vocabulary: 3.0/5.0, Phraseology: 3.0/5.0, Grammar: 4.0/5.0, Conventions: 3.0/5.0}
cohesion	syntax	vocabulary	phraseology	grammar	conventions	
    
    \item \textbf{Rationale (GPT-3.5-Turbo):} \textbf{[conventions]:} "they wont have to change and get up early in the morning to shower and do there hair." - Spelling and grammar errors. "most students usually take showers before school." - Spelling errors and lack of punctuation. \textbf{[grammar]:} "they wont have to change and get up early in the morning to shower and do there hair." (Grammar errors: "wont" should be "won\'t," "there" should be "their"). \textbf{[phraseology]:} "get up early in the morning to shower and do there hair." - Incorrect grammar and spelling. "you wake up go brush your teeth and go to your closet and look at your cloths." - Incorrect grammar and spelling. "most students usually take showers before school." - Relevant to the topic but lacks variety in phrases.\textbf{[vocabulary]:} "they wont have to change and get up early in the morning to shower and do there hair." - Incorrect word usage and spelling errors. "most students usually take showers before school." - Simple vocabulary, lacks complexity. \textbf{[syntax]:} "The hardest part of school is getting ready." - Well-written, simple sentence. "most students usually take showers before school." - Minor errors in sentence structure. "when your home your comfortable and you pay attention." - Basic errors in sentence structure. \textbf{[cohesion]:} "taking only classes helps them because at there house they\'ll be pay more attention." (Poorly written - lacks proper grammar and cohesion) "most students usually take showers before school." (Fairly written - lacks cohesion but clear in meaning) "when your home your comfortable and you pay attention." (Poorly written - lacks proper grammar and cohesion)
    
    \item \textbf{Rationale (Llama-3.1-8B-Instruct):}
    \textbf{[cohesion]:} The essay demonstrates a limited range of cohesive devices, primarily relying on repetition of ideas and phrases to connect sentences and paragraphs. There is a lack of transitional words and phrases, resulting in abrupt shifts between ideas, and some sentences seem disconnected from the rest of the text. \textbf{[syntax]:} The essay demonstrates a limited range of syntactic structures, primarily using simple sentences with some compound structures. There are occasional errors in sentence formation, such as incorrect subject-verb agreement ("they wont have to change") and missing articles ("at there house"). \textbf{[vocabulary]:} The essay demonstrates a narrow range of vocabulary, relying on basic words and phrases. Topic-related terms are occasionally used, but often inappropriately. There are errors in word formation and choice, such as "do there hair" and "pay more attention," which distort meanings. \textbf{[phraseology]:} The essay demonstrates a narrow range of phraseological devices, primarily relying on simple collocations and lexical bundles, such as "take classes," "get up early," and "pay attention." Repetitions and misuses of phrases are evident, with an over-reliance on basic sentence structures. \textbf{[grammar]:} The essay demonstrates inconsistent grammar usage, with several errors in verb tense, subject-verb agreement, and punctuation. Incorrect word order, missing articles, and misuse of pronouns are also prevalent. However, the writer attempts to convey their ideas and uses basic sentence structures effectively. \textbf{[conventions]:} The essay demonstrates a developing use of conventions to convey meaning. While there are some errors in spelling (e.g., "there" instead of "their"), capitalization (e.g., missing capitalization in sentences), and punctuation (e.g., missing periods), the overall structure and syntax are mostly coherent.
\end{itemize}

\end{document}

%% file: abstract.tex
Existing automated essay scoring (AES) has solely relied on essay text without using explanatory rationales for the scores, thereby forgoing an opportunity to capture the specific aspects evaluated by rubric indicators in a fine-grained manner. This paper introduces Rationale-based Multiple Trait Scoring (RMTS), a novel approach for multi-trait essay scoring that integrates prompt-engineering-based large language models (LLMs) with a fine-tuning-based essay scoring model using a smaller large language model (S-LLM). RMTS uses an LLM-based trait-wise rationale generation system where a separate LLM agent generates trait-specific rationales based on rubric guidelines, which the scoring model uses to accurately predict multi-trait scores. Extensive experiments on benchmark datasets, including ASAP, ASAP++, and Feedback Prize, show that RMTS significantly outperforms state-of-the-art models and vanilla S-LLMs in trait-specific scoring. By assisting quantitative assessment with fine-grained qualitative rationales, RMTS enhances the trait-wise reliability, providing partial explanations about essays. The code is available at \textbf{\url{https://github.com/BBeeChu/RMTS.git}}.

%% file: intro.tex
\begin{figure}[h]
    \centering
    \includegraphics[trim=0 0 0 0, clip, width=\linewidth]{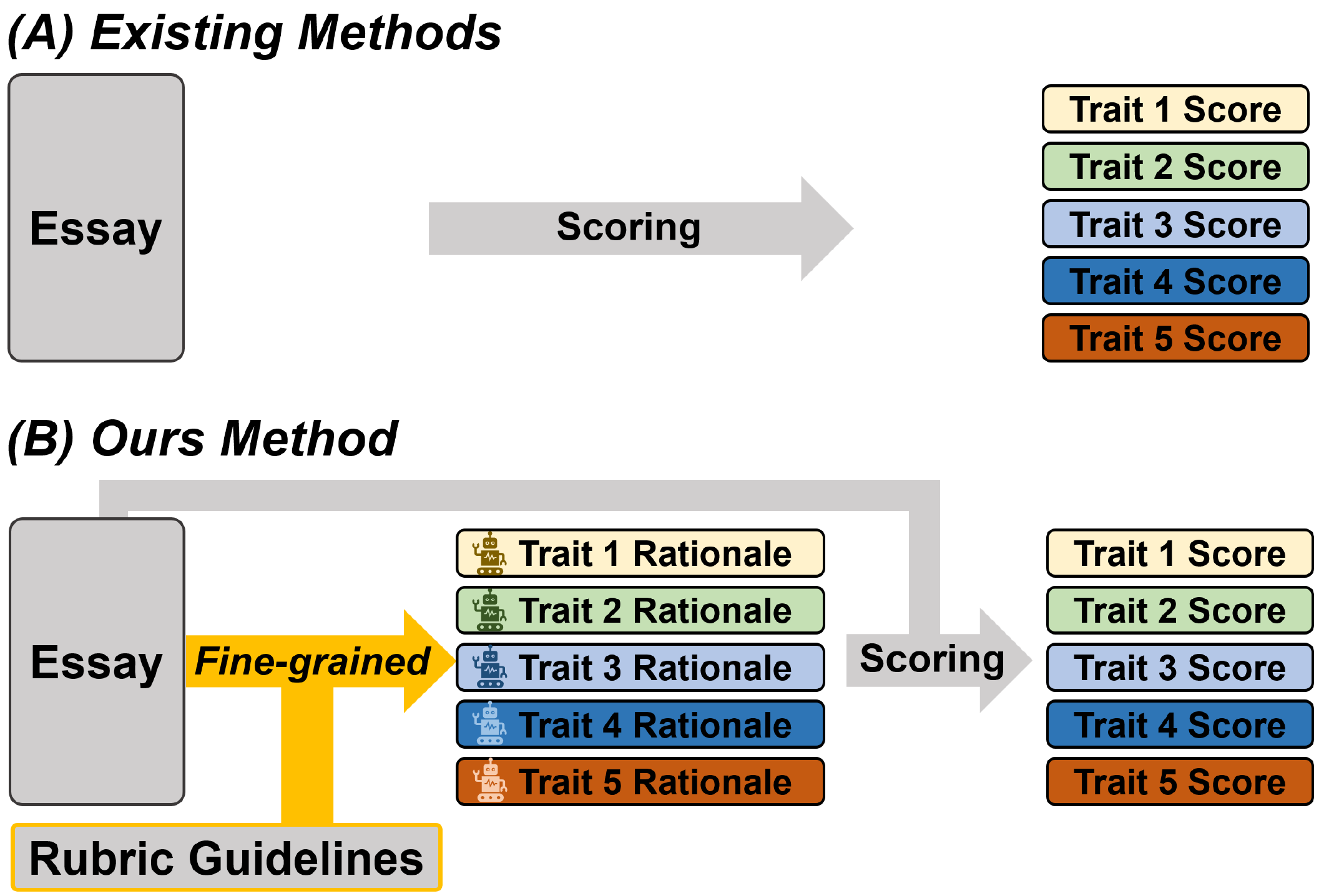}
    \caption{Unlike existing methods (A), we use multiple prompt-engineering LLMs to generate trait-specific rationales based on rubric guidelines as shown in (B), which are then combined with an S-LLM for comprehensive evaluation.}
    \label{fig1}
\end{figure}

Multi-trait essay scoring, which evaluates essays on multiple dimensions such as \textit{Content}, \textit{Organization}, and \textit{Style}, rather than on a single holistic score, has recently become a central issue in automated essay scoring (AES). Extensive research in this area has primarily utilized BERT and trait-wise layers to predict scores for individual traits \cite{mathias, ridley, kumar, do}. Notably, \citet{do2024} proposed using an autoregressive pre-trained language model, T5 \cite{raffel}, for greater computational efficiency. Despite these efforts, most studies have used essay texts alone to predict labels as represented in Figure \ref{fig1} (A), rather than extracting aspects evaluated by rubric indicators from the essays and using them.

With the advent of LLMs, generating fine-grained rationales—explanations of how essays align with rubric criteria—has become feasible. As shown in Figure \ref{fig1}, incorporating rationales identifies relevant essay sections that demonstrate specific traits and links them directly to the rubric. This approach mirrors how human evaluators use rubrics to assess essays in real-world settings \cite{freeman}. For instance, a rationale for \textit{Organization} highlights transitions and structure, leading to more precise, rubric-aligned evaluations. Without rationales, the model may overlook key elements and score less accurately by focusing only on the semantic sequence.

To the best of our knowledge, few studies have attempted to use rationale-based evaluations derived from rubrics to assess essays \cite{lee, li}. \citet{lee} used LLMs to predict holistic scores based on criteria synthesized from rubrics, but the models cannot be fine-tuned as this approach relies on prompt engineering. \citet{li} combined LLM-generated rationales with human-assessed scores to create new labels for training sequence-to-sequence models. However, this method does not use the rationales as inputs to the encoders when predicting scores. Besides, it differs from our research as the work only focused on predicting overall score of short-answer responses rather than multiple trait scores of long-context essays.

In this paper, we propose a novel approach of effectively utilizing rationales in conjunction with rubics and essays to enable LLMs to better assess the various aspects of essays as outlined by the rubrics. These rationales, or qualitative assessments, are then used by an encoder-decoder-based smaller large language model, hereafter referred to as S-LLM, to predict scores more precisely. The S-LLM is fine-tuned using training data that includes human-rated scores. This approach aims to enhance the accuracy and reliability of automated multi-trait essay scoring.

The main contributions of this study are as follows:
\begin{itemize}
    \item We introduce a novel approach to multi-trait scoring, \textbf{R}ationale-based \textbf{M}ulti-\textbf{T}rait essay \textbf{S}coring (\textbf{RMTS}), which combines an essay and a rationale together to predict multi-trait scores. This model utilizes the rationale to explicitly capture the elements assessed by the rubric from the essay.
    \item We compose trait-specific prompts using essays and rubrics to build an LLM-based trait-wise rationale generation system. This system generates rationales, which serve as the foundation for the multi-trait scores.
    \item We conducted a comprehensive analysis of the generated rationales and verified that they are sufficiently meaningful to be effectively utilized in essay scoring.
    \item Extensive experiments with five S-LLMs demonstrate that incorporating LLM-generated rationales significantly improves essay scoring, providing a model-agnostic incremental benefit to each S-LLM. Our approach advances the state-of-the-art baselines in essay scoring on the ASAP and ASAP++ benchmark datasets.
\end{itemize}

%% file: related_work.tex
\subsection{Traditional and transformer-based automated essay scoring}
Traditional automated essay scoring (AES) focused on holistic scoring, predicting an overall score using handcrafted features and linear regression models \cite{taghipour, dong2016, dong2017, cozma}. Particularly, transformer-based models like BERT \cite{devlin} significantly improved AES by capturing detailed language information \cite{yang, wang, mayfield2020should}. These models enhanced scoring accuracy but were primarily used for holistic scoring. Extending them to multi-trait scoring is inefficient due to the need for multiple models for different traits, increasing computational costs \cite{kumar, do2024}.

\subsection{Multi-trait essay scoring approaches}
Multi-trait AES evaluates essays across various dimensions with respect to different features existing in essays such as \textit{Content}, \textit{Organization}, and \textit{Conventions}. Existing models used multiple linear layers or separate models for each trait, which require intensive resources \cite{mathias,ridley}. Recent approaches introduced multi-task learning frameworks with shared models and trait-specific layers, improving efficiency \cite{kumar}. However, handling trait dependencies and requiring specialized modules remains challenging. The autoregressive multi-trait scoring (ArTS) model addressed this by using a pre-trained T5 model to sequentially generate trait scores, leveraging inter-dependencies for better accuracy \cite{do2024}. Yet, it still relied solely on essay texts alone for score prediction.

\subsection{Rubric-based essay scoring using large language model}

Recently, LLMs have been used to evaluate essays alongside assessment rubrics, showing competitive performance. For example, one study \cite{lee} divided criteria into multiple traits and generated sub-criteria for scoring, achieving moderate results. Another study \cite{li} used rubrics to score short answers and generated rationales, which were then used as labels to fine-tune the T5 model \cite{raffel} to produce both scores and rationales. However, this approach did not outperform fine-tuned models like BERT \cite{devlin} and Longformer \cite{beltagy}, which were trained using only scores as labels. Our approach differs by directly extracting rubric-based rationales from essays using LLMs and feeding them into a pre-trained S-LLM. This method explicitly considers detailed scoring criteria, improving alignment with human evaluators and enhancing both the reliability and transparency of automated essay scoring.

%% file: rmts.tex
\begin{figure*}[h]
    \centering
    \includegraphics[width=0.85\linewidth]{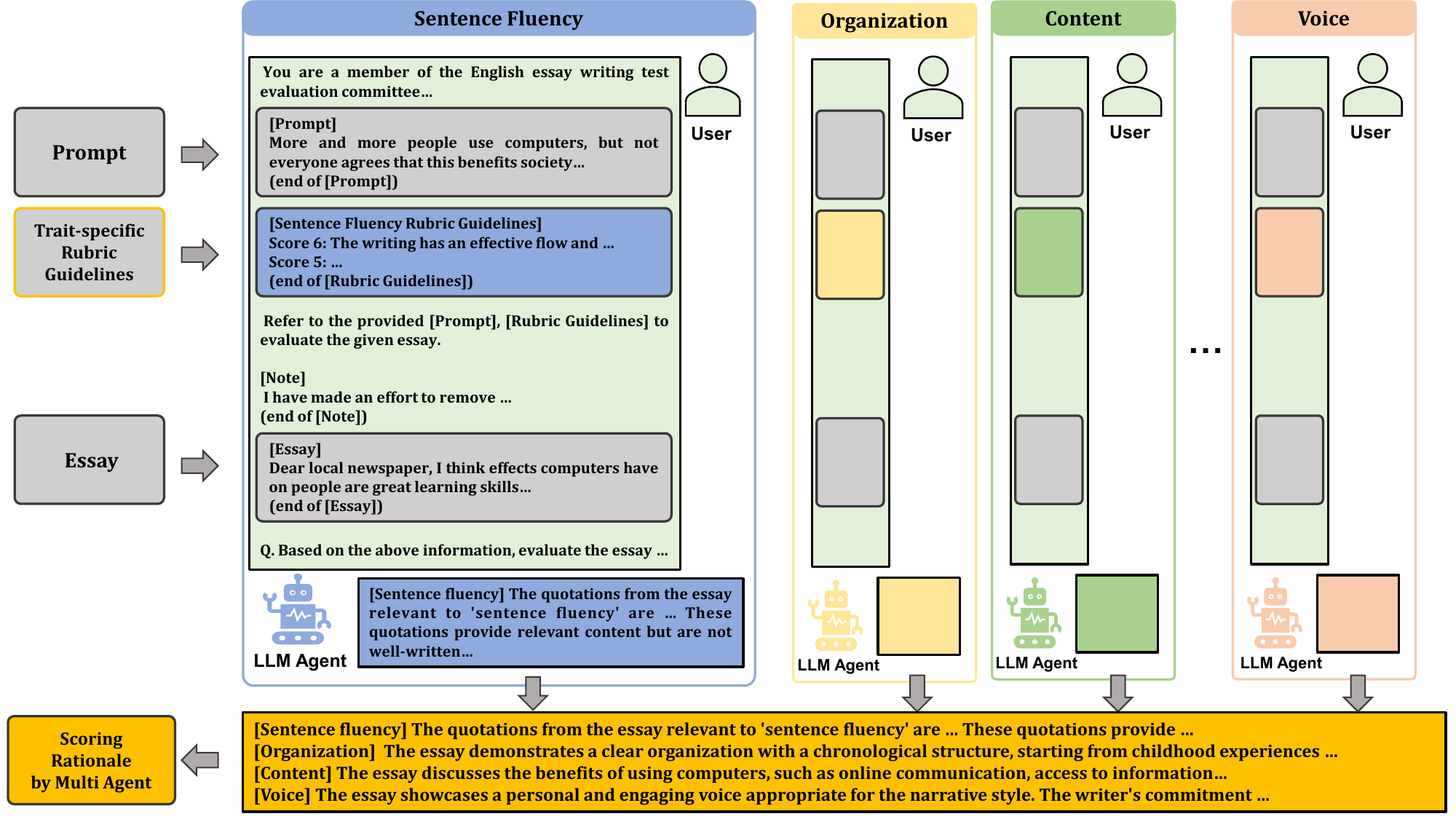}
    \caption{Trait-specific rationales are constructed using the essay prompt, the essay, and the rubric guidelines corresponding to each trait. To generate the final rationale for each essay, we combine the trait-specific rationales in sequence.}
    \label{fig2}
\end{figure*}

\begin{figure}[h!]
    \centering
    \includegraphics[trim=0 60 0 60, clip, width=\linewidth]{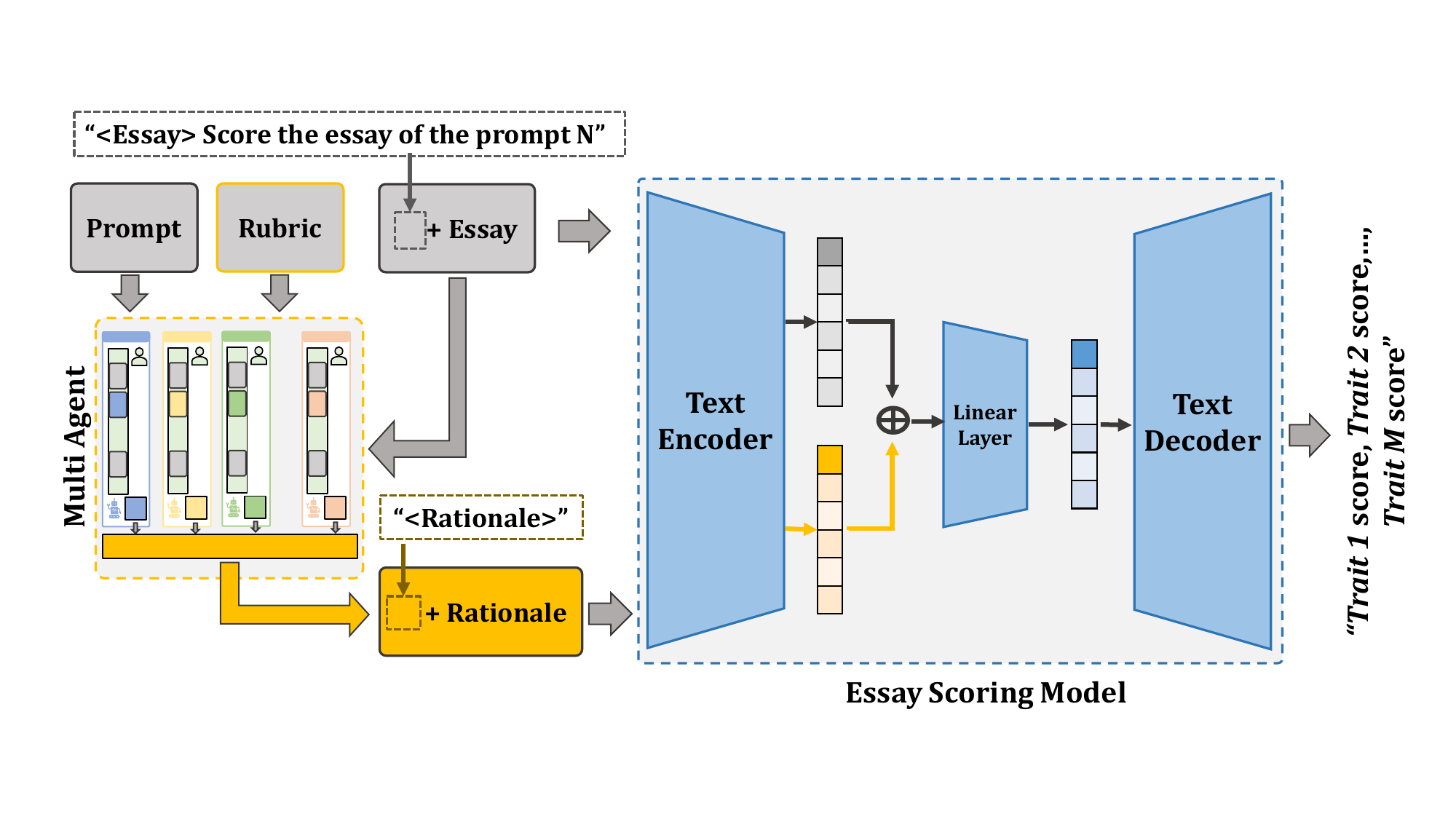}
    \caption{The final rationale generated by multiple LLM agents and the essay are fed into a shared encoder to extract their representations. These representations are then projected to a unified feature vector by a linear layer and passed through the decoder, which predicts trait-specific scores in sequence.}
    \label{fig3}
\end{figure}

RMTS is a framework that enhances the multi-trait essay scoring capabilities of an S-LLM, a pre-trained sequence-to-sequence model, by incorporating rationales. The framework consists of two parts: (1) generating trait-specific rationales using an LLM-based system with GPT-3.5 Turbo \cite{chatgpt}\footnote{https://platform.openai.com/docs/models} and Llama-3.1-8B-Instruct \cite{touvron}\footnote{https://llama.meta.com/responsible-use-guide/} (referred to as GPT and Llama respectively), and (2) extracting representations from both the essay and rationale using a shared encoder of the S-LLM. This dual-process approach improves the reliability of the scoring model. The detailed procedure is shown in Figures \ref{fig2} and \ref{fig3}.

\subsection{LLM-based trait-wise rationale generation system}

As illustrated in Figure \ref{fig2}, individual trait-specific prompts are constructed using the essay and the rubric corresponding to each trait. Each trait-specific prompt is then provided to a separate LLM agent dedicated to that trait. This approach, referred to as the LLM-based trait-wise rationale generation system, relies on the LLM's demonstrated ability to effectively evaluate essays, as supported by prior research \cite{lee, ho, li}. We have adopted the prompts used in \citet{lee} as a basis for the task description and modified them to fit our context. We have also added trait-specific rubric to them. Our trait-wise LLM agents generate qualitative assessments based on the rubric, producing rationales in a text form. This method enables the generation of detailed, text-based rationales that are directly tied to the rubric, facilitating a subsequent S-LLM to decide the final numeric score in a more accurate manner.

Given that the decoder of the S-LLM used in RMTS predicts subsequent tokens based on previous ones, the rationale was also constructed in a sequence that evaluates sub-dimensional, constituent traits of the essay first (e.g., \textit{Content}, \textit{Organization}, \textit{Style}), followed by the overall trait. This approach capitalizes on the model's capability to boost predictive performance by replicating the sequential nature of human assessors when evaluating traits \cite{do2024}. 

\subsection{Representation extraction and scoring}
In the current study, we utilize various pre-trained encoder-decoder S-LLMs for scoring multi-traits of essays. We include five widely used models—T5, Flan-T5, BART, Pegasus, and LED (Longformer Encoder-Decoder model) \cite{raffel, flant5, lewis, pegasus, led}—as the S-LLMs for essay scoring. Figure \ref{fig3} shows the RMTS architecture. Each component in RMTS framework’s essay scoring model corresponds to the respective component of the individual S-LLM. 

In RMTS, both the essay and the generated rationales are fed into a single encoder to extract their respective representations, which means that the two texts share a common encoder, allowing their representations to be projected into the same vector space. Inspired by \cite{do2024}, we add the prompt \textit{"Score the essay of the prompt N"} to the essay text to improve model inference. Special tokens, such as "\textit{<Essay>}" and "\textit{<Rationale>}," are inserted before the essay and rationale to help the tokenizer distinguish between the two. We also introduce tokens for multi-trait names (e.g., \textit{<Content>}) to prevent them from being split into sub-words, preserving their meanings.

The encoder processes this combined input to generate dense representations, which are integrated into a unified feature vector by a linear layer for scoring. This vector is passed to a decoder, which predicts trait-specific scores. By leveraging both the essay and rationale, the model delivers detailed multi-trait scoring. 

\subsection{Score extraction}

Since we use S-LLMs, which are sequence-to-sequence models, we predict and generate scores for multiple traits alongside their respective names from each essay one at a time, based on techniques from \citet{do2024}. The generated string of scores is transformed into a dictionary format, where trait names serve as keys to extract the scores. For accurate evaluation, we disregard predictions for traits whose ground truth values are \textit{NaN}.

%% file: experiment.tex
In this study, we conducted extensive experiments to analyze the generated rationales and evaluate their effectiveness in scoring multiple essay traits, guided by the following research questions.

\begin{itemize}
    \item \textbf{RQ1.} What are the key findings from the analysis of LLM-generated rationales for essay evaluation?
    \item \textbf{RQ2.} To what extent does incorporating rationales improve the reliability of multi-trait essay scoring using S-LLMs?
\end{itemize}

\subsection{Datasets} \label{datasets}

In our main experiment, we utilized the ASAP\footnote{https://www.kaggle.com/competitions/asap-aes/data} and ASAP++\footnote{https://lwsam.github.io/ASAP++/lrec2018.html} \cite{mathias} datasets, comprising English essays from American high school students (grades 7–10) across eight prompts. The ASAP dataset provides overall scores for all essays, but only prompts 7 and 8 have trait-specific scores. Thus, we included ASAP++ for rated trait scores on the remaining prompts, and this combined dataset will be referred to as "ASAP/ASAP++" throughout the paper. Additionally, the Feedback Prize dataset\footnote{https://www.kaggle.com/competitions/feedback-prize-english-language-learning/data}, which consists of argumentative essays written by American students (grades 6–12) and labeled with six traits, was used without distinguishing between prompts to examine the generalizability of the incremental effect of using essays and rationales together on vanilla S-LLMs. Due to space constraints, the dataset descriptions are provided in Table \ref{dataset_description}.

\begin{table}[!htbp]
\centering
\scriptsize
\setlength{\tabcolsep}{2pt}
\begin{tabular}{>{\centering\arraybackslash}m{1cm}|c|c|p{4cm}}
    \toprule
    \textbf{Dataset} & \textbf{Prompt} & \textbf{\# Essays} & \textbf{Traits} \\
    \midrule
    \multirow{8}{*}{\rotatebox{90}{\textbf{ASAP/ ASAP++}}} 
    & 1 & 1785 & Over, Cont, WC, Org, SF, Conv \\
    & 2 & 1800 & Over, Cont, WC, Org, SF, Conv \\
    & 3 & 1726 & Over, Cont, PA, Nar, Lang \\
    & 4 & 1772 & Over, Cont, PA, Nar, Lang \\
    & 5 & 1805 & Over, Cont, PA, Nar, Lang \\
    & 6 & 1800 & Over, Cont, PA, Nar, Lang \\ 
    & 7 & 1569 & Over, Cont, Org, Conv, Style \\
    & 8 & 723 & Over, Cont, WC, Org, SF, Conv, Voice \\
    \midrule
    \multirow{1}{*}{\textbf{Feedback}}
    & - & 3930 & Coh, Syn, Voca, Phr, Gram, Conv \\
    \bottomrule
\end{tabular}
\caption{Composition of the ASAP/ASAP++ combined dataset, listing writing traits per prompt. Traits include: Over: \textit{Overall}, Cont: \textit{Content}, WC: \textit{Word Choice}, Org: \textit{Organization}, SF: \textit{Sentence Fluency}, Conv: \textit{Conventions}, PA: \textit{Prompt Adherence}, Nar: \textit{Narrativity}, Lang: \textit{Language}, with Feedback Prize traits being: Coh: \textit{Cohesion}, Syn: \textit{Syntax}, Voca: \textit{Vocabulary}, Phr: \textit{Phraseology}, Gram: \textit{Grammar}, Conv: \textit{Conventions}.}
\label{dataset_description}
\end{table}

\subsection{Rationale Analysis}
To evaluate rationale quality, we performed various analyses. We evaluated the similarity of the generated rationales using ROUGE-L \cite{rouge} on a sample of 100 essays to analyze the diversity in how LLMs generate them.  Additionally, we measured the faithfulness of LLM-generated rationales to the predicted multi-trait scores, using a proxy method from prior studies \cite{wiegreffe2020, jain2020, li}. Specifically, we fine-tuned S-LLMs to predict multi-trait scores using only the rationales as input to the models.
 
\subsection{Baselines}
To compare performance across the two datasets described earlier, we used five widely adopted vanilla S-LLMs, all encoder-decoder models \cite{raffel, flant5, lewis, pegasus, led} designed for text generation tasks. We also included baseline models with a string kernel based model and RNN-based architectures from the referenced papers: HISK \cite{cozma}, STL-LSTM \cite{dong2017}, MTL-BiLSTM \cite{kumar}, PMAES \cite{chen2023pmaes}, and PLAES \cite{chen2024plaes}. These models align with our main task of multi-trait scoring (see Appendix \ref{baseline_detail} for details on each baseline). For a fair comparison with the traditional benchmark datasets (ASAP/ASAP++), we used the performance data of four baseline models—HISK \cite{cozma}, STL-LSTM \cite{dong2017}, MTL-BiLSTM \cite{kumar}, and ArTS \cite{do2024}—as reported in \cite{do2024}, along with two additional baseline models, PMAES \cite{chen2023pmaes} and PLAES \cite{chen2024plaes}, as reported in their original papers. ArTS is a model that employs the vanilla T5-base model for scoring multi-trait essays.

\subsection{Experimental Settings}
In this study, we employed GPT-3.5-Turbo\footnote{https://openai.com/index/openai-api/} and Llama-3.1-8B-Instruct\footnote{https://llama.meta.com/responsible-use-guide/} for rationale generation based on a prompt-engineering technique, and fine-tuned pre-trained S-LLMs from Huggingface\footnote{https://huggingface.co/}. Using the Seq2SeqTrainer from the same platform, models were trained over 15 epochs with a batch size of 4, and evaluations took place every 5000 steps, applying early stopping with a patience of 2. All experiments were conducted on a single NVIDIA A100 GPU using the PyTorch framework.

\subsection{Evaluation}

To ensure consistent evaluation, we utilized 5-fold cross-validation across all models, employing a 60/20/20 split for training, validation, and testing, following the methodology of \citet{taghipour} and \citet{kumar} with the combined ASAP and ASAP++ dataset. For the Feedback Prize dataset, we applied the same 5-fold process but with stratified splitting based on label distribution. Assessment was conducted using quadratic weighted kappa (QWK) \cite{cohen}, the dataset’s designated metric, which effectively measures score disparities between human raters and model predictions. We chose the top two models from each fold and reported the highest QWK as the final result \cite{do2024}.

%% file: results.tex
\subsection{Rationale Analysis (RQ1)}
We focus on analyzing the rationales from the ASAP/ASAP++ and Feedback Prize datasets in terms of similarity and faithfulness.

\begin{figure}[htbp]
    \centering
    \includegraphics[width=\linewidth]{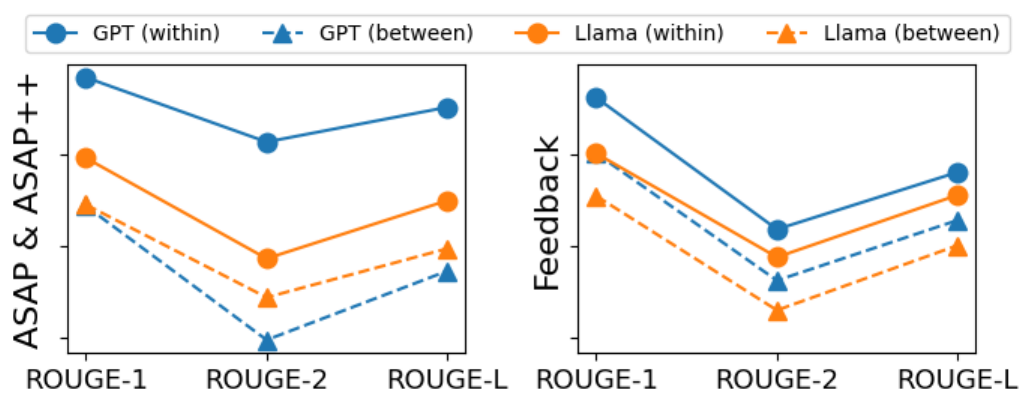}
    \caption{ROUGE scores of rationales \textit{within} the same essay or \textit{between} different essays across GPT and Llama.}
    \label{rouge_asap_fig}
\end{figure}

\begin{table*}[h!]
\centering
\scriptsize
\setlength{\tabcolsep}{0.5pt} 
\label{table2}
\resizebox{\textwidth}{!}{%
\begin{tabular}{l|l|l|l|l|l|l|l|l|l|l|l|l}
    \toprule
     & \multicolumn{11}{c|}{\textbf{Trait (Prediction Order: $\leftarrow$) }}  \\
    \midrule
     \textbf{Model} &\textbf{Overall} & \textbf{Cont} & \textbf{PA} & \textbf{Lang} & \textbf{Nar} & \textbf{Org} & \textbf{Conv} & \textbf{WC} & \textbf{SF} & \textbf{Style} & \textbf{Voice} & \textbf{AVG$\uparrow$ (SD$\downarrow$)} \\
    \midrule
    HISK &0.718 &0.679 &0.697 &0.605 &0.659 &0.610 &0.527 &0.579 &0.553 &0.609 &0.489 &0.611 (0.004) \\
    STL-LSTM & 0.750 & 0.707 & 0.731 & 0.640 & 0.699 & 0.649 & 0.505 & 0.621 & 0.612 & 0.609 & 0.544 & 0.642 (0.073) \\
    MTL-BiLSTM & \textbf{0.764} & 0.685 & 0.701 & 0.604 & 0.668 & 0.615 & 0.560 & 0.615 & 0.598 & 0.632 & 0.582 & 0.639 (0.057) \\
    PMAES & 0.671 & 0.567 &  0.584 &  0.545 &  0.614 & 0.481 &  0.421 &  0.584 &  0.582 &- &- & 0.614 (-) \\
    PLAES & 0.673 & 0.574 & 0.601 & 0.554 &  0.631 & 0.491 & 0.447  & 0.579 &0.580 &- &- & 0.631 (-) \\
    \midrule
    T5 (ArTS) & 0.754 & 0.730 & \underline{0.751} & 0.698 & 0.725 & 0.672 & 0.668 & 0.679 & 0.678 & \textbf{0.721} & 0.570 & 0.695 (0.018) \\
    + RMTS(G) \tiny{(+\%)}& \underline{0.755} \tiny{(+0.1)} & \textbf{0.737} \tiny{(+0.7)} & \textbf{0.752} \tiny{(+0.1)} & \textbf{0.713} \tiny{(+1.5)} & \textbf{0.744} \tiny{(+1.9)} & \underline{0.682} \tiny{(+1.0)} & \textbf{0.690} \tiny{(+2.2)} & \textbf{0.705} \tiny{(+2.6)} & \textbf{0.694} \tiny{(+1.6)} & \underline{0.702} \tiny{(-1.9)} & 0.612 \tiny{(+4.2)} & \textbf{0.708} (0.043) \\
    + RMTS(L) \tiny{(+\%)} & 0.754 \tiny{(+0.0)} &	0.730 \tiny{(+0.0)} &	0.749 \tiny{(-0.2)} &	0.701 \tiny{(+0.3)} &	\underline{0.737} \tiny{(+1.2)} &	0.675 \tiny{(+0.3)} &	\underline{0.684} \tiny{(+1.6)} &	0.690 \tiny{(+1.1)} &	\underline{0.684} \tiny{(+0.6)} &	0.696 \tiny{(-2.5)} &	\underline{0.640} \tiny{(+7.0)} &	\underline{0.704} (0.042)\\

    \midrule
    Flan-T5 & 0.662 & 0.645 & 0.615 & 0.539 & 0.577 & 0.646 & 0.636 & \underline{0.694} & 0.667 & 0.578 & 0.624 & 0.626 (0.064) \\
    + RMTS(G) \tiny{(+\%)}& 0.732 \tiny{(+7.0)} & \underline{0.733} \tiny{(+8.8)} & 0.750 \tiny{(+13.5)} & \underline{0.708} \tiny{(+16.9)} & \underline{0.737} \tiny{(+16.0)} & \textbf{0.684} \tiny{(+3.8)} & 0.680 \tiny{(+4.4)} & 0.691 \tiny{(-0.3)} & 0.680 \tiny{(+1.3)} & 0.688 \tiny{(+11.0)} & 0.563 \tiny{(-6.1)} & 0.695 (0.048) \\

    + RMTS(L) \tiny{(+\%)}& 0.723 \tiny{(+6.1)} &	0.717 \tiny{(+7.2)} &	0.736 \tiny{(+12.1)} &	0.696 \tiny{(+15.7)} &	0.722 \tiny{(+14.5)} &	0.663 \tiny{(+1.7)} &	0.662 \tiny{(+2.6)} &	0.673 \tiny{(-2.1)} &	0.663 \tiny{(-0.4)} &	0.695 \tiny{(+11.7)} &	0.620 \tiny{(-0.4)} &	0.688 (0.054)\\

    \midrule
    BART & 0.701 & 0.672 & 0.711 & 0.664 & 0.705 & 0.600 & 0.588 & 0.624 & 0.601 & 0.646 & 0.547 & 0.642 (0.054) \\
    + RMTS(G) \tiny{(+\%)}& 0.720 \tiny{(+1.9)} & 0.710 \tiny{(+3.8)} & 0.731 \tiny{(+2.0)} & 0.683 \tiny{(+1.9)} & 0.720 \tiny{(+1.5)} & 0.651 \tiny{(+5.1)} & 0.637 \tiny{(+4.9)} & 0.685 \tiny{(+6.1)} & 0.655 \tiny{(+5.4)} & 0.661 \tiny{(+1.5)} & \textbf{0.649} \tiny{(+10.2)} & 0.674 (0.046) \\
    + RMTS(L)  \tiny{(+\%)}& 0.724 \tiny{(+2.3)} & 0.704 \tiny{(+3.2)} & 0.732 \tiny{(+2.1)} & 0.677 \tiny{(+1.3)} & 0.714 \tiny{(+0.9)} & 0.658 \tiny{(+5.8)} & 0.647 \tiny{(+5.9)} & 0.671 \tiny{(+4.7)} & 0.662 \tiny{(+6.1)} & 0.673 \tiny{(+2.7)} & 0.596 \tiny{(+4.9)} & 0.678 (0.037) \\

    \midrule
    Pegasus & 0.536 & 0.584 & 0.608 & 0.586 & 0.629 & 0.578 & 0.515 & 0.559 & 0.519 & 0.578 & 0.388 & 0.553 (0.065) \\
    + RMTS(G) \tiny{(+\%)} & 0.711 \tiny{(+17.5)} & 0.651 \tiny{(+6.7)} & 0.698 \tiny{(+9.0)} & 0.674 \tiny{(+8.8)} & 0.697 \tiny{(+6.8)} & 0.615 \tiny{(+3.7)} & 0.600 \tiny{(+8.5)} & 0.618 \tiny{(+5.9)} & 0.613 \tiny{(+9.4)} & 0.619 \tiny{(+4.1)} & 0.561 \tiny{(+17.3)} & 0.641 (0.046) \\
    + RMTS(L) \tiny{(+\%)} & 0.713 \tiny{(+17.7)} & 0.650 \tiny{(+6.6)} & 0.698 \tiny{(+9.0)} & 0.667 \tiny{(+8.1)} & 0.699 \tiny{(+7.0)} & 0.624 \tiny{(+4.6)} & 0.605 \tiny{(+9.0)} & 0.640 \tiny{(+8.1)} & 0.626 \tiny{(+10.7)} & 0.638 \tiny{(+6.0)} & 0.570 \tiny{(+18.2)} & 0.648 (0.041) \\

    \midrule
    LED & 0.709 & 0.677 & 0.706 & 0.666 & 0.707 & 0.627 & 0.633 & 0.643 & 0.640 & 0.655 & 0.522 & 0.653 (0.053) \\
    + RMTS(G) \tiny{(+\%)} & 0.736 \tiny{(+2.7)} & 0.714 \tiny{(+3.7)} & 0.733 \tiny{(+2.7)} & 0.688 \tiny{(+2.2)} & 0.719 \tiny{(+1.2)} & 0.667 \tiny{(+4.0)} & 0.663 \tiny{(+3.0)} & 0.676 \tiny{(+3.3)} & 0.674 \tiny{(+3.4)} & 0.694 \tiny{(+3.9)} & 0.597 \tiny{(+7.5)} & 0.687 (0.038) \\
    + RMTS(L) \tiny{(+\%)} & 0.727 \tiny{(+1.8)} & 0.711 \tiny{(+3.4)} & 0.741 \tiny{(+3.5)} & 0.674 \tiny{(+0.8)} & 0.714 \tiny{(+0.7)} & 0.656 \tiny{(+2.9)} & 0.648 \tiny{(+1.5)} & 0.658 \tiny{(+1.5)} & 0.644 \tiny{(+0.4)} & 0.684 \tiny{(+2.9)} & 0.542 \tiny{(+2.0)} & 0.673 (0.052)\\

    \bottomrule
\end{tabular}
}
\caption{Average QWK scores across all prompts for each trait on the ASAP/ASAP++ datasets. The values in parentheses (\%) represent the percentage of improvement in RMTS performance when incorporating rationales generated by GPT (G) or Llama (L) compared to the vanilla S-LLMs. Traits are predicted from right to left 
 ($\leftarrow$), and five-fold averaged standard deviation is reported (SD). The best results are highlighted in \textbf{bold}, and the second-best results are \underline{underlined}. For PMAES and PLAES, style and voice traits, where contrastive learning is not feasible due to their presence in a single prompt, are marked as "-".}
\label{asap_trait_performance}
\end{table*}

\subsubsection{Similarity of rationales}

Figure \ref{rouge_asap_fig} displays the similarity analysis results for rationales generated by two LLMs, presenting ROUGE scores for the ASAP/ASAP++ and Feedback Prize datasets. To evaluate consistency, we calculated ROUGE scores between rationales generated for the same essay across five iterations and averaged them, labeled as \textit{"within"}. In RMTS, we used the first rationale generated from the five iterations. To gauge diversity, we computed ROUGE scores between the rationales used in RMTS for different essays, also averaged and labeled as \textit{"between"}. As shown, \textit{"within"} scores are higher than \textit{"between"} scores, indicating each LLM captures and reflects the unique characteristics of each essay in its rationale while maintaining consistent features within the same essay.


\subsubsection{Faithfulness of rationales}

\begin{figure}[h!]
    \centering
    \includegraphics[width=\linewidth]{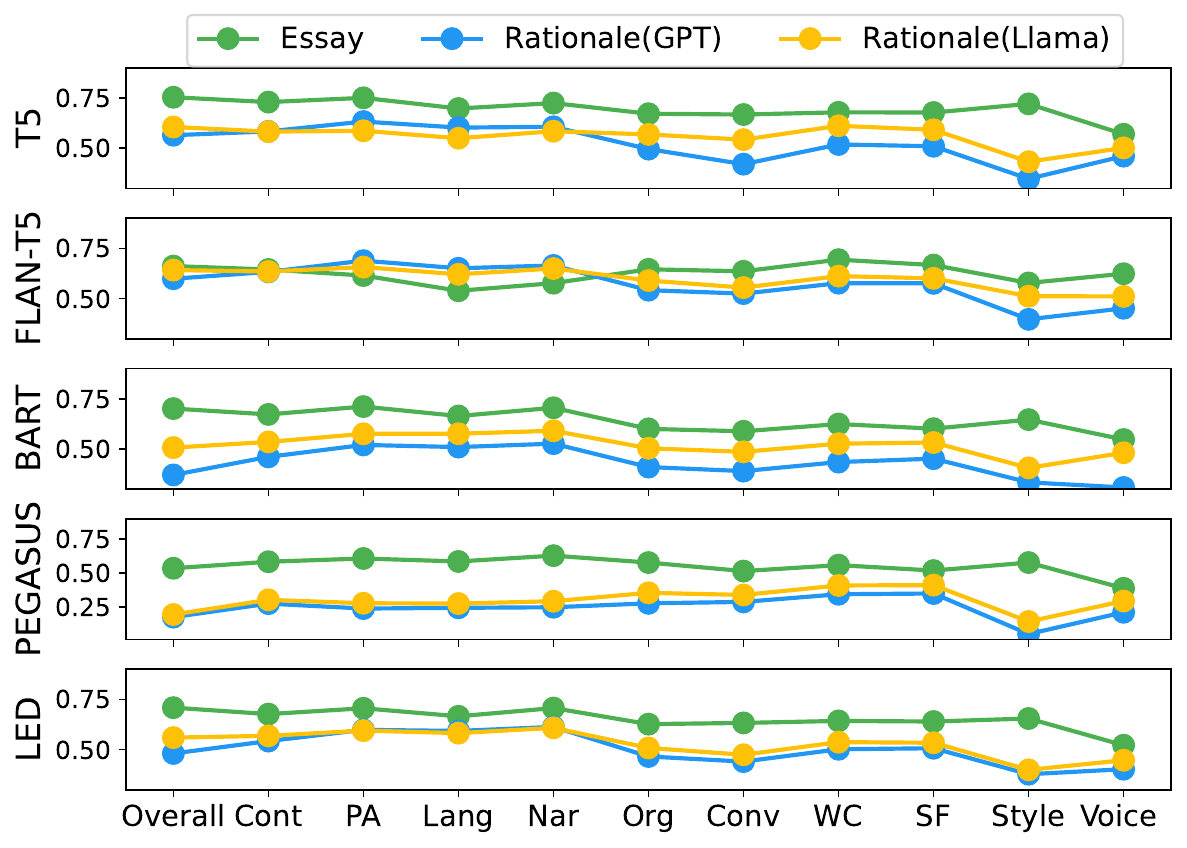} 
    \caption{Performance comparison of S-LLMs based on QWK scores, averaged across all prompts for each trait with regard to the ASAP/ASAP++ dataset, using either the essays or the rationales generated by GPT or Llama.}
    \label{combined_trait_fig}
\end{figure}

Figures \ref{combined_trait_fig} compares the reliability (measured by QWK against human-labeled scores) of each model in predicting essay scores for the ASAP/ASAP++ dataset, using either the essays or the LLM-generated rationales. Most models performed at over 80\% of their essay-only performance in nearly entire traits when using the rationales, demonstrating that rationales make a meaningful contribution to S-LLMs' essay evaluations. Given that these rationales are qualitative free-text outputs, this also indicates that they provide partial explanations for the models' score predictions \cite{wiegreffe2020, jain2020, li}. However, as shown in Figure \ref{combined_trait_fig}, Pegasus achieved about 40\% of its essay-only performance with GPT-generated rationales and 50\% with Llama-generated rationales in average, suggesting that it relies more on the intrinsic features of essays than on qualitative evaluation data.

\begin{table*}[h!]
\centering
\scriptsize
\setlength{\tabcolsep}{1pt} 
\label{table2}
\begin{tabular}{l|l|l|l|l|l|l|l|l|l}
    \toprule
     & \multicolumn{8}{c|}{\textbf{Prompt}}  \\
    \midrule
    \textbf{Model} & \textbf{1} & \textbf{2} & \textbf{3} & \textbf{4} & \textbf{5} & \textbf{6} & \textbf{7} & \textbf{8} & \textbf{AVG$\uparrow$(SD$\downarrow$)} \\
    \midrule
    HISK &0.674 &0.586 &0.651 &0.681 &0.693 &0.709 &0.641 &0.516 &0.644 (0.004) \\
    STL-LSTM & 0.690 & 0.622 & 0.663 & 0.729 & 0.719 & 0.753 & 0.704 & 0.592 & 0.684 (0.055) \\
    MTL-BiLSTM & 0.670 & 0.611 & 0.647 & 0.708 & 0.704 & 0.712 & 0.684 & 0.581 & 0.665 (0.048) \\
    PMAES & 0.656 & 0.553 & 0.598 & 0.606 & 0.626 & 0.572 & 0.386 & 0.530 & 0.566 (-) \\
    PLAES & 0.648 & 0.563 & 0.604 & 0.623 & 0.634 & 0.593 & 0.403 & 0.533 & 0.575 (-) \\
    \midrule

    T5 (ArTS) & 0.708 & \textbf{0.706} & 0.704 & 0.767 & 0.723 & \textbf{0.776} & \textbf{0.749} & 0.603 & 0.717 (0.025) \\
    + RMTS(G)\tiny{(+\%)} & \textbf{0.716}\tiny{(+0.8)} & \underline{0.704}\tiny{(-0.2)} & \textbf{0.723}\tiny{(+1.9)} & \textbf{0.772}\tiny{(+0.5)} & \textbf{0.737}\tiny{(+1.4)} & 0.769\tiny{(-0.7)} & \underline{0.736}\tiny{(-1.3)} & \underline{0.651}\tiny{(+4.8)} & \textbf{0.726} (0.042) \\
    + RMTS(L)\tiny{(+\%)} & 0.705\tiny{(-0.3)} & 0.692\tiny{(-1.4)} & \underline{0.714}\tiny{(+1.0)} & 0.766\tiny{(-0.1)} & 0.726\tiny{(+0.3)} & \underline{0.773}\tiny{(-0.3)} & 0.726\tiny{(-2.3)} & \textbf{0.658}\tiny{(+5.5)} & \underline{0.720} (0.044) \\

    \midrule
    Flan-T5 & 0.703 & 0.691 & 0.523 & 0.599 & 0.593 & 0.674 & 0.609 & 0.633 & 0.628 (0.056) \\

    + RMTS(G)\tiny{(+\%)} & \underline{0.711}\tiny{(0.8)} & 0.666\tiny{(-2.5)} & \textbf{0.723}\tiny{(+20.0)} & \underline{0.771}\tiny{(+17.2)} & \underline{0.736}\tiny{(+14.3)} & 0.762\tiny{(+8.8)} & 0.723\tiny{(+11.4)} & 0.642\tiny{(0.9)} & 0.717 (0.055) \\

    + RMTS(L)\tiny{(+\%)} & 0.700\tiny{(-0.3)} & 0.643\tiny{(-4.8)} & 0.702\tiny{(+17.9)} & 0.761\tiny{(+16.2)} & 0.719\tiny{(+12.6)} & 0.751\tiny{(+7.7)} & 0.734\tiny{(+12.5)} & 0.623\tiny{(-0.1)} & 0.704 (0.055) \\

    \midrule
    BART & 0.647 & 0.602 & 0.658 & 0.727 & 0.713 & 0.713 & 0.624 & 0.534 & 0.652 (0.066) \\
    + RMTS(G)\tiny{(+\%)} & 0.707\tiny{(+6.0)} & 0.667\tiny{(+6.5)} & 0.702\tiny{(+4.4)} & 0.751\tiny{(+2.4)} & 0.718\tiny{(+0.5)} & 0.737\tiny{(+2.4)} & 0.684\tiny{(+6.0)} & 0.595\tiny{(+6.1)} & 0.695 (0.045) \\
    + RMTS(L)\tiny{(+\%)} & 0.698\tiny{(+5.1)} & 0.658\tiny{(+5.6)} & 0.691\tiny{(+3.3)} & 0.744\tiny{(+1.7)} & 0.720\tiny{(+0.7)} & 0.748\tiny{(+3.5)} & 0.690\tiny{(+6.6)} & 0.614\tiny{(+8.0)} & 0.695 (0.042) \\

    \midrule
    Pegasus & 0.639 & 0.520 & 0.518 & 0.562 & 0.636 & 0.597 & 0.539 & 0.478 & 0.561 (0.058) \\
    + RMTS(G)\tiny{(+\%)} & 0.672\tiny{(+3.3)} & 0.631\tiny{(+11.1)} & 0.683\tiny{(+16.5)} & 0.725\tiny{(+16.3)} & 0.718\tiny{(+8.2)} & 0.695\tiny{(+9.8)} & 0.593\tiny{(+5.4)} & 0.573\tiny{(+9.5)} & 0.661 (0.057) \\
    + RMTS(L)\tiny{(+\%)} & 0.670\tiny{(+3.1)} & 0.637\tiny{(+11.7)} & 0.679\tiny{(+16.1)} & 0.714\tiny{(+15.2)} & 0.708\tiny{(+7.2)} & 0.714\tiny{(+11.7)} & 0.611\tiny{(+7.2)} & 0.587\tiny{(+10.9)} & 0.665 (0.046) \\

    \midrule
    LED & 0.704 & 0.650 & 0.679 & 0.705 & 0.701 & 0.707 & 0.638 & 0.520 & 0.663 (0.064) \\
    + RMTS(G)\tiny{(+\%)} & 0.701\tiny{(-0.3)} & 0.684\tiny{(+3.4)} & 0.693\tiny{(+1.4)} & 0.762\tiny{(+5.7)} & 0.721\tiny{(+2.0)} & 0.742\tiny{(+3.5)} & 0.715\tiny{(+7.7)} & 0.620\tiny{(+10.0)} & 0.705 (0.004) \\
    + RMTS(L)\tiny{(+\%)} & 0.694\tiny{(-1.0)} & 0.654\tiny{(+0.4)} & 0.688\tiny{(+0.9)} & 0.754\tiny{(+4.9)} & 0.724\tiny{(+2.3)} & 0.745\tiny{(+3.8)} & 0.714\tiny{(+7.6)} & 0.592\tiny{(+7.2)} & 0.696 (0.049) \\

    \bottomrule
\end{tabular}
\caption{Average QWK scores across all traits for each prompt on the ASAP/ASAP++ datasets. The values in parentheses (\%) represent the percentage of improvement in RMTS performance when incorporating rationales generated by GPT (G) or Llama (L) compared to the vanilla S-LLMs. Five-fold averaged standard deviation is reported (SD). The best results are highlighted in \textbf{bold}, and the second-best results are \underline{underlined}.}
\label{asap_prompt_performance}
\end{table*}

\subsection{Performance Comparison (RQ2)}
To address the second research question, we first evaluate RMTS against the baseline models and vanilla S-LLMs on the ASAP/ASAP++ dataset, the standard for essay scoring. We then assess RMTS against vanilla S-LLMs on the Feedback Prize dataset to demonstrate its broader applicability. Finally, to gain deeper insights into the role of rationales, we fine-tuned three S-LLMs with rationales and essays, removing one trait at a time during the process to observe performance variations.

\subsubsection{Performance with ASAP/ASAP++}

Since our target is to predict individual trait scores, we will focus on trait scoring rather than \textit{Overall} scores. Owing to space constraints, the results are presented in Table \ref{asap_trait_performance} and \ref{asap_prompt_performance}. Table \ref{asap_trait_performance} shows model performance on the ASAP/ASAP++ dataset. Using GPT-generated rationales, RMTS applied to each of the five S-LLMs outperforms their respective vanilla versions across nearly all traits, except for \textit{Style} in T5 and \textit{Word Choice} and \textit{Voice} in Flan-T5. T5 model shows incremental improvements with rationales, ranking first or second in every trait, including \textit{Overall}. Additionally, RMTS with T5, BART, and LED outperforms the best traditional models—HISK, STL-LSTM, and MTL-BiLSTM—in every trait using GPT rationales. Although MTL-BiLSTM has a higher \textit{Overall} score, the gap with RMTS-T5 is small, and RMTS focuses more on individual trait scoring (Additional performance comparisons with baselines specifically designed for overall score assessment are provided in Appendix \ref{overall_comparison}).  

On top of that, Table \ref{asap_prompt_performance} shows prompt-wise performance. RMTS improves the vanilla S-LLMs for the majority of prompts. With GPT-generated rationales, RMTS using T5, Flan-T5, and LED generally outperforms their vanilla counterparts across most prompts, and other S-LLMs show improvements with the same rationales in every prompt. Additionally, RMTS with T5 and Flan-T5 using GPT rationales outperforms the best traditional baseline models.

\subsubsection{Performance with Feedback Prize Dataset}
\label{feedback_performance_section}
To assess the broader applicability of using rationales to enhance S-LLMs in essay scoring, we conducted additional experiments with the Feedback Prize dataset, as shown in Table \ref{feedback_performance}. Despite the small dataset size of 2.3K samples—about one-third of the ASAP/ASAP++ dataset—rationales improve the performance of the four S-LLMs (T5, BART, Pegasus, and LED) across most traits. However, integrating rationales does not improve the vanilla Flan-T5 model. We attribute this to the model's inherent characteristics from instruction-fine-tuning \cite{flant5}, which may prevent it from effectively incorporating rationales with such a small dataset. Nevertheless, these results indicate that rationales generally enhance model performance, even in data-scarce environments.

\begin{table}[h!]
\centering
\scriptsize
\setlength{\tabcolsep}{1.2pt}

\resizebox{\columnwidth}{!}{%
\begin{tabular}{l|l|l|l|l|l|l|l}
    \toprule
     & \multicolumn{6}{c|}{\textbf{Trait (Prediction Order: $\leftarrow$)}}  \\
    \midrule
    \textbf{Model} & \textbf{Conv} & \textbf{Gram} & \textbf{Phr} & \textbf{Voc} & \textbf{Syn} & \textbf{Coh} &\textbf{AVG$\downarrow$ (SD$\uparrow$)}  \\
    \midrule
    HISK & 0.279 & 0.279 & 0.270 & 0.261 & 0.252 & 0.241 & 0.264(0.012) \\
    STL & 0.544 & 0.440 & 0.534 & 0.515 & 0.518 & 0.459 & 0.502(0.024) \\
    MTL-BiLSTM & 0.527 & 0.484 & 0.505 & 0.519 & 0.507 & 0.462 & 0.501(0.027) \\
   \midrule
    T5      &   0.521 &	0.479 &	0.512 &	0.454 &	0.497 &	0.467 &	0.488(0.027) \\
    + RMTS(G) \tiny{(+\%)}  & 0.568	\tiny{(+4.7)}&0.550	\tiny{(+7.1)}&\underline{0.543}	\tiny{(+3.1)}&0.430	\tiny{(-2.4)}&\underline{0.543}	\tiny{(+4.6)}&0.498	\tiny{(+3.1)}&0.522 (0.024)	 \\
    
    + RMTS(L) \tiny{(+\%)} & \underline{0.570} \tiny{(+4.9)} & \textbf{0.557} \tiny{(+7.8)} & 0.535 \tiny{(+2.3)} & 0.443 \tiny{(-1.1)} & 0.534 \tiny{(+3.7)} & 0.490 \tiny{(+2.3)} & 0.522 (0.024) \\

    \midrule
    Flan-T5 & 0.539&	0.512&	0.527&	\textbf{0.466}	&0.531	&0.491&	0.511 (0.025) \\
    + RMTS(G) \tiny{(+\%)} & 0.520\tiny{(-1.9)} &	0.507\tiny{(-0.5)} &	0.497\tiny{(-3.0)} &	0.440\tiny{(-2.6)} &	0.513\tiny{(-1.8)} &	0.472\tiny{(-1.9)} &	0.492	(0.034) \\
    + RMTS(L) \tiny{(+\%)} & 0.479\tiny{(-6.0)}&	0.476\tiny{(-3.6)}&	0.513\tiny{(-1.4)}&	0.407\tiny{(-5.9)}&	0.496\tiny{(-3.5)}&	0.477\tiny{(-1.4)}&	0.475 (0.123)\\

    \midrule
    BART       & 0.396 & 0.357 & 0.440 & 0.363 & 0.288 & 0.314 & 0.360 (0.050) \\
    + RMTS(G) \tiny{(+\%)}  & 0.565 \tiny{(+16.9)} & 0.477 \tiny{(+12.0)} & \textbf{0.596} \tiny{(+15.6)} & 0.461 \tiny{(+9.8)} & 0.507 \tiny{(+21.9)} & \textbf{0.573} \tiny{(+25.9)} & \underline{0.530} (0.051) \\
    + RMTS(L) \tiny{(+\%)} & 0.439 \tiny{(+4.3)} & 0.410 \tiny{(+5.3)} & 0.366 \tiny{(-7.4)} & 0.329 \tiny{(-3.4)} & 0.341 \tiny{(+5.3)} & 0.172 \tiny{(-14.2)} & 0.343 (0.085) \\

    \midrule
    Pegasus    & 0.273 & 0.233 & 0.265 & 0.304 & 0.270 & 0.264 & 0.268 (0.021) \\
    + RMTS(G) \tiny{(+\%)} & 0.290 \tiny{(+1.7)} & 0.237 \tiny{(+0.4)} & 0.309 \tiny{(+4.4)} & 0.315 \tiny{(+1.1)} & 0.313 \tiny{(+4.3)} & 0.334 \tiny{(+7.0)} & 0.299 (0.031) \\
    + RMTS(L) \tiny{(+\%)} & 0.327 \tiny{(+5.4)} & 0.273 \tiny{(+4.0)} & 0.359 \tiny{(+9.4)} & 0.363 \tiny{(+5.9)} & 0.350 \tiny{(+8.0)} & 0.369 \tiny{(+10.5)} & 0.340 (0.033) \\

    \midrule
    LED        & 0.520 & 0.479 & 0.486 & 0.428 & 0.505 & 0.476 & 0.482 (0.029) \\
    + RMTS(G) \tiny{(+\%)} & \textbf{0.586} \tiny{(+6.6)} & \underline{0.552} \tiny{(+7.3)} & 0.540 \tiny{(+5.4)} & \underline{0.462} \tiny{(+3.4)} & \textbf{0.550} \tiny{(+4.5)} & \underline{0.505} \tiny{(+2.9)} & \textbf{0.533} (0.039) \\
    + RMTS(L) \tiny{(+\%)}  & 0.565 \tiny{(+4.5)} & 0.531 \tiny{(+5.2)} & 0.507 \tiny{(+2.1)} & 0.428 \tiny{(+0.0)} & 0.520 \tiny{(+1.5)} & 0.461 \tiny{(-1.5)} & 0.502 (0.045) \\

    \bottomrule
\end{tabular}%
}
\caption{Average QWK scores across all prompts for each trait on the Feedback Prize dataset. The values in parentheses (\%) represent the percentage of improvement in RMTS performance when incorporating rationales generated by GPT (G) or Llama (L) compared to the vanilla S-LLMs. Traits are predicted from right to left 
 ($\leftarrow$), and five-fold averaged standard deviation is reported (SD). The best results are highlighted in \textbf{bold}, and the second-best results are \underline{underlined}. PMAES and PLAES, which utilize prompt-wise contrastive learning, were excluded as they cannot be trained on datasets where all essays share a single prompt.}
 \label{feedback_performance}
\end{table}

\subsection{Trait Rationale Ablation Study (RQ2)}
\label{asap_ablation_section}
To assess the effectiveness of trait-specific rationales, we conducted an ablation study by removing a trait commonly shared across prompts, as illustrated in Figure \ref{fig4}. We focused on traits present in at least four prompts. Specifically, \textit{Content} is evaluated in all prompts, while \textit{Organization} and \textit{Conventions} are scored in Prompts 1, 2, 7, and 8. \textit{Prompt Adherence}, \textit{Language}, and \textit{Narrativity} are assessed in Prompts 3 to 6. For comparison, we divided the prompts into two groups and excluded \textit{Word Choice}, \textit{Sentence Fluency}, \textit{Style}, and \textit{Voice} since they appear in fewer prompts. We used T5, Flan-T5, and BART in the experiments, fine-tuning each model by removing one trait at a time.

A consistent decline in performance is observed across all traits when the corresponding rationale is removed, confirming that trait rationales significantly influence their respective assessments. For instance, removing the rationale for \textit{Conventions} results in a performance drop for that trait, particularly when compared to RMTS, which utilizes rationales for all traits. Although the performance of models without a trait rationale lag behind RMTS, it still outperforms vanilla models without any rationale input. This suggests that trait rationales not only influence their own assessments but also interact with and affect the evaluation of other traits \cite{canale}. For example, when the rationale for \textit{Conventions} is removed, performance still surpass that of vanilla models.

Interestingly, performance does not always drop significantly when certain trait rationales are removed, particularly in the case of T5 when \textit{Organization} and \textit{Prompt Adherence} rationales are excluded. This implies that the effectiveness of rationales can vary by trait and model.

Overall, the results show that rationales generally have an incremental effect on performance. Essay scoring performs worse when trait rationales are removed, highlighting the crucial role these rationales play in predicting trait scores.

\begin{figure}[h!]
    \centering
    \label{fig4}
    \includegraphics[width=\linewidth]{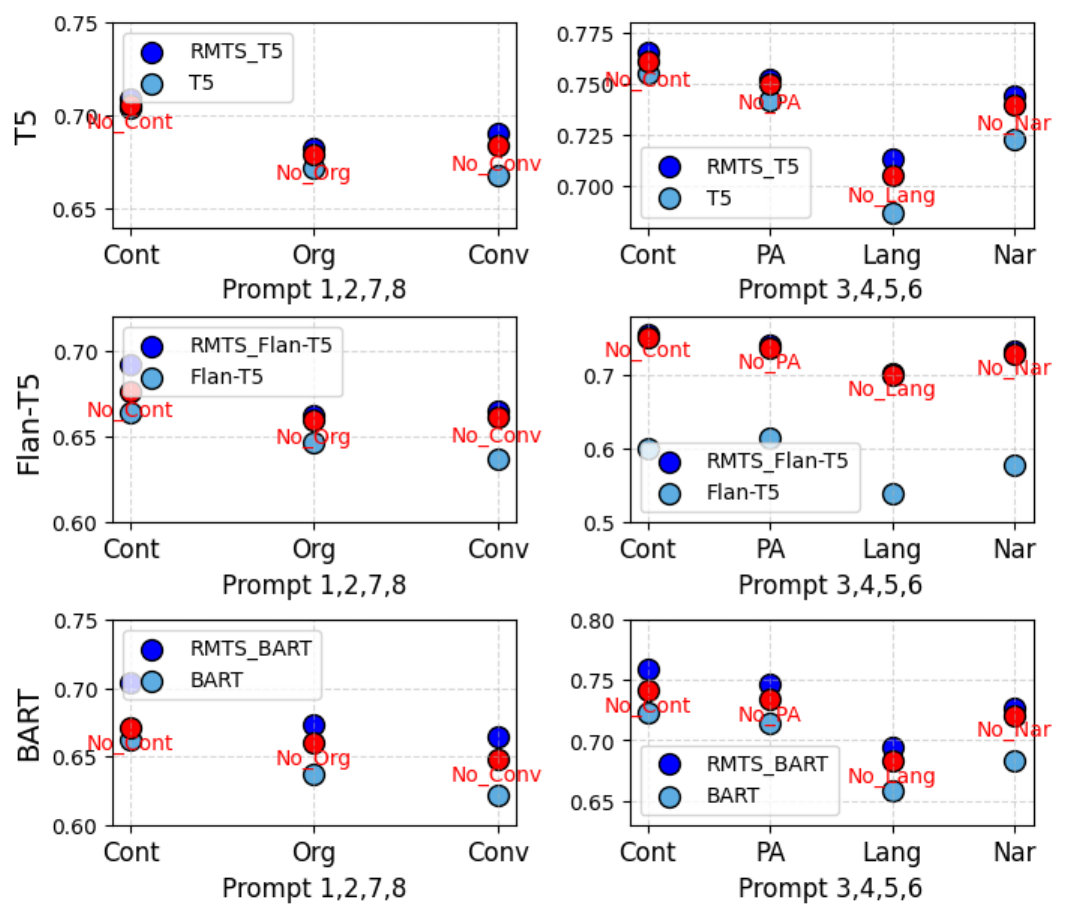} 
    \caption{Ablation study of rationale when removing each of the trait in \textbf{(A) Prompt 1,2,7,8} and \textbf{(B) Prompt 3,4,5,6}. The performances generally drop when any one trait is omitted, underscoring the importance of incorporating all traits in rationale generation.}
    \label{fig4}
\end{figure}

%% file: conclusion.tex
This paper introduces RMTS, a framework that uses prompt-engineering-based LLMs to improve multi-trait essay scoring in S-LLMs by generating trait-specific rationales aligned with rubric guidelines and incorporating them into the scoring process. Our results show that RMTS with S-LLMs significantly improves the performance of each vanilla model, with RMTS using T5 even outperforming state-of-the-art baselines. Additionally, removing rationales negatively impacts performance. These study findings highlight the substantial benefits of utilizing trait-specific rationales generated by LLMs, which has been untapped by prior research. From this view point, RMTS can be seen as opening up new horizons for automated essay scoring with S-LLMs.